\documentclass[10pt,journal,cspaper,compsoc,onecolumn]{elsarticle}

\usepackage{fullpage}
\usepackage{enumitem}

\usepackage{epsfig}
\usepackage{graphicx}
\usepackage{amsmath}
\usepackage{amssymb}
\usepackage{caption}
\usepackage{subcaption}
\usepackage{booktabs}
\usepackage[table]{xcolor}
\usepackage{multirow}
\usepackage{algorithm}
\usepackage{algorithmic}
\usepackage{dsfont}

\usepackage{float}
\usepackage{soul}

\def\eg{{\em e.g.,~}}

\begin{document}

\title{Online Algorithms for Factorization-Based Structure from Motion}%

\author[add1]{Ryan Kennedy\corref{mycorrespondingauthor}}
\cortext[mycorrespondingauthor]{Corresponding author at: Department of Computer and Information Science,
University of Pennsylvania, Philadelphia PA}
\ead{kenry@cis.upenn.edu}
\author[add2]{Laura Balzano}
\author[add3]{Stephen J.~Wright}
\author[add1]{Camillo J.~Taylor}

\address[add1]{University of Pennsylvania}
\address[add2]{University of Michigan}
\address[add3]{University of Wisconsin-Madison}
%\author{Ryan~Kennedy,
%        Laura~Balzano,
%        Stephen~J.~Wright,
%        and~Camillo~J.~Taylor% <-this % stops a space
%\IEEEcompsocitemizethanks{\IEEEcompsocthanksitem R.~Kennedy and C.~J.~Taylor are with
%the University of Pennsylvania.
%E-mail: \{kenry,cjtaylor\}@cis.upenn.edu
%\IEEEcompsocthanksitem L. Balzano is with the University of Michigan.
%Email: girasole@umich.edu
%\IEEEcompsocthanksitem S. Wright is with the University of Wisconsin-Madison.
%Email: swright@cs.wisc.edu}
%\thanks{}}

\markboth{November 2014}{}%

\begin{abstract}
%\boldmath
  We present a family of online algorithms for real-time factorization-based
  structure from motion, leveraging a relationship between the incremental singular value
  decomposition and recently proposed methods for online matrix completion. Our
  methods are orders of magnitude faster than previous state of the art, 
  can handle missing data and a variable number
  of feature points, and are robust to noise and sparse outliers. 
  We demonstrate our methods on both real and synthetic sequences and show
  that they perform well in both online and batch settings. We also 
  provide an implementation that is able to produce 3D
  models in real time using a laptop with a webcam.
\end{abstract}

% Note that keywords are not normally used for peer review papers.


\maketitle

\section{Introduction}

The problem of structure from motion --- recovering the 3D structure
of an object and locations of the camera from a monocular video stream
--- has been studied extensively in computer vision.
For the rigid case, which is the focus of this paper, many algorithms are based on the seminal work of
Tomasi and Kanade~\cite{tomasi1992shape}. In this work it was shown that a
noise-free measurement matrix of point tracks has rank at most 3 for
an affine camera when the data are centered at the origin. The 3D locations of all tracked points and camera
positions can be easily obtained from a factorization of this
matrix. Due to occlusion, however, the matrix is typically missing
many entries,  so standard matrix factorization techniques cannot
be applied.

Recent work in {\em low-rank matrix completion} has explored conditions under
which the missing entries in a low-rank matrix can be determined, even
when the matrix is corrupted with noise or sparse
outliers~\cite{candestao,RechtImprovedMC09,CandesRPCA09,candesplan}. Most
algorithms for matrix completion are static in nature, working with a
``batch'' of matrix columns. In this paper, 
% by contrast,
 we focus on
\emph{online} structure from motion, in which the 3D model of point
locations must be updated in real time as the video is captured.  The
algorithm must be efficient enough to run in real time, yet still must
deal effectively with missing data and noisy observations.  The online
problem has received little attention in comparison to batch
algorithms, although online algorithms have been developed for
matrix completion~\cite{balzano_grouse, grouseconvergence} that have shown
promise in other real-time computer vision applications~\cite{he2012cvpr}. In this
paper, we extend these online algorithms to the problem of rigid
structure from motion.

Our main contribution is a suite of online matrix completion algorithms that can be applied to the
structure from motion (SFM) problem. The main algorithm SAGE is based on~\cite{balzano_grouse}, but differs fundamentally in that it is based on matrix factorization instead of incremental gradient descent. Our other algorithms are variations on SAGE, and they inherit this property.
Our algorithms address several difficulties
that have been observed in 
this field, specifically:  (1) our
method is inherently online and provides a matrix estimate after every new update, (2) we
are able to handle a dynamically changing number of features,
(3) we directly deal with missing data, (4) we naturally deal with data that are
offset from the origin, (5) our
algorithms can be made robust to outliers, and (6) our method is
extremely fast.  We note that (1) and (2) together imply that
  our algorithms can accommodate changes in the both the numbers of
  rows and columns.

In addition to testing with online data, we 
% compare our methods to
% batch algorithms, showing
show that our algorithms are
 competitive with -- and often orders of magnitude faster than -- 
state-of-the-art batch algorithms. To demonstrate the utility of our approach,
we describe a laptop implementation 
% of our algorithm that is able to
that creates 3D models of objects in real time using
video from an attached webcam.

% We don't need to mention this but can include it in the letter to
% the editor
% This paper builds on our previous conference publication \cite{kennedy2013}. We include
% an extended description of our algorithm, a new discussion section and  more thorough
% experiments.

\section{Related Work}

\subsection{Structure From Motion}

Much research on the rigid structure-from-motion problem is based on
Tomasi and Kanade's factorization algorithm~\cite{tomasi1992shape} for
orthographic cameras, and subsequent
work~\cite{poelman1997paraperspective} that extended its applicability
to other camera
models~\cite{aanaes2002robust,buchanan2005damped,gotardo2011computing,guerreiro20023d,morita1997sequential,tardif2007algorithms}.
Comparatively little work has been done on \emph{online} structure
from motion, apart from algorithms that employ batch methods or local
bundle adjustment in an online
framework~\cite{klein2007parallel,mouragnon2009generic}.

In \cite{morita1997sequential}, Mortia and Kanade proposed a
sequential version of the factorization algorithm, but their method cannot deal with
missing data, nor can they handle outliers or a dynamically changing
set of features.  The approach of McLauchlan and
Murray~\cite{mclauchlan1995unifying} can handle missing data but uses
simplifying heuristics to achieve a low computational complexity. The
related algorithm of Trajkovi\'{c} and
Hedley~\cite{trajkovic1997practical} dispenses with the heuristics, but  focuses on tracking of moving objects within a scene.
More recently, Cabral et al.~\cite{cabral2011fast} proposed a method
that performs matrix completion iteratively, in an online manner.
However, it is difficult to  add new features dynamically with their
approach, and it  can be numerically unstable.
The algorithm most similar to our own is the incremental SVD (ISVD) approach
of Bunch and Nielsen~\cite{Bunch78}, which has been previously adapted to handle missing
data~\cite{brand2002incremental, grouseisvd}.  In Section~\ref{sec:isvd} we show how 
ISVD is related to our algorithms.

\subsection{Matrix Completion and Subspace Tracking}

Low-rank matrix completion is the problem of 
recovering a low-rank matrix from an incomplete sample of the entries.
It was shown in \cite{RechtImprovedMC09,candesrecht} that under
assumptions on the number of observed entries and
on incoherence of the singular vectors of this matrix with respect to the canonical coordinate axes, the
nuclear norm minimization convex optimization problem solves the
NP-hard rank minimization problem exactly. Since this breakthrough, a flurry
of research activity has centered around developing faster algorithms
to solve this convex optimization problem, both exactly and
approximately; see~\cite{Keshavan10b,Toh09} for two examples.  The algorithm
GROUSE~\cite{balzano_grouse} (Grassmannian Rank-One Update Subspace
Estimation) performs incremental gradient descent on a non-convex version of the matrix completion
problem that admits fast online updates. GROUSE outperforms all non-parallel 
algorithms in computational efficiency, often by an order of magnitude, while remaining
competitive in terms of estimation error when the noise is small.

The GROUSE algorithm was developed for low-dimensional subspace tracking with incomplete data.
This is an area of extensive research. Comprehensive reference lists for complete-data adaptive methods for
tracking subspaces and extreme singular values and vectors of
covariance matrices can be found in~\cite{balzanophd,Edelman98}, where 
methods from the matrix computation literature 
and gradient-based methods from the signal processing literature are all discussed.
Since GROUSE is an online method, it is a natural candidate for structure from motion, but it is not quite adequate in that it estimates only the {\em column space} of the matrix, and requires a final projection of all incomplete columns onto this column space to complete the matrix. Therefore, the unadorned form of  GROUSE is not appropriate for contexts where (1) the number of rows is changing dynamically and (2) real-time completion is of interest --- both of which are true in real-time structure from motion. Our algorithm SAGE (named after the sage grouse) addresses both these issues using a matrix factorization formulation.

A useful extension of these matrix completion algorithms is toward ``Robust PCA,'' 
which seeks to recover a low-rank matrix in the presence of
outliers \cite{strelow2012general,eriksson2010efficient,wang2012probabilistic,zheng2012practical}. This work has had application in computer vision,
decomposing a scene from several video frames as the sum of a low-rank
matrix of background (which represents the global appearance and
illumination of the scene) and a sparse matrix of moving foreground
objects \cite{CandesRPCA09,mateos10}. 
% The algorithm
GRASTA~\cite{he2012cvpr,he2011grasta} (Grassmannian Robust Adaptive Subspace Tracking
Algorithm) is a robust extension of
GROUSE that performs online estimation of a changing low-rank subspace
while subtracting outliers. Our algorithm RSAGE is a robust version of SAGE that features a similar extension.

\subsection{Contributions of SAGE}

We now comment on the novelty of SAGE with respect to previous contributions.
While the original ISVD algorithm provides a matrix estimate that is updated with each 
streaming column or added row, it does not naturally handle missing data
or revisiting past points. While the GROUSE algorithm handles missing data, 
it does not update the row space simultaneously with the column space, so does not provide
a full matrix estimate at each iteration.
SAGE combines the benefits of the two approaches: It allows missing data and
 provides a matrix estimate at every iteration. Further, SAGE inherits the speed of GROUSE, making
it orders of magnitude faster than most other algorithms on a 
variety of structure from motion problems, and allowing it to be used in truly real-time contexts.

The version of ISVD that is adapted to missing
  data~\cite{brand2002incremental, grouseisvd} has a drawback: It
  relies heavily on early subspace estimates, since the update
  incorporates singular value estimates. Therefore when there is a lot
  of missing data, early estimates of the subspace and singular values
  are poor and convergence is slow. By contrast, the GROUSE algorithm
  is agnostic to previous data; all directions of the current subspace
  carry equal weight. SAGE inherits this property of GROUSE.

Minor contributions of SAGE include the incorporation of the all-ones
offset vector into the column space, the ability to revisit previous
columns to improve the matrix estimate, and a variant that is robust
to sparse outliers. Additionally, it is novel among matrix completion
algorithms in its ability to handle added rows, to the best of our
knowledge.  This feature makes it potentially useful in several other
applications, \eg recommender systems, where both the number of users
and number of products are changing over time.

In summary, this paper presents the first fully online factorization-based
structure-from-motion algorithms, whose speed improvements over
earlier techniques allow them to be applied in real-time
settings.

\section{Problem Description}
\label{sec:sfmdescription}

Given a set of points tracked through a video, the \emph{measurement
  matrix} $W$ is defined as
\begin{equation}
W = \begin{bmatrix}x_{1,1}&\dots&x_{1,m}\\\vdots&\ddots&\vdots\\x_{n,1}&\dots&x_{n,m}\end{bmatrix},
\end{equation}
where $x_{i,j}\in\mathbb{R}^{1\times2}$ is the projection of point $i$ onto the camera at frame
$j$, giving $W$ a size of $n \times 2m$, where $n$ is the number of points and $m$ is the number of frames. If we assume that the data are centered at the origin, then 
in the absence of noise
this matrix has rank at most $3$ and can be factored as
\begin{equation}
W = \tilde{S}\tilde{M}^T,
\end{equation}
where $\tilde{S}$ is the $n\times3$ \emph{structure} matrix containing 3D
point locations, while $\tilde{M}$ is the $2m\times 3$ \emph{motion} matrix,
which is the set of affine camera matrices corresponding to each
frame~\cite{tomasi1992shape}. Data that are offset from the origin have an additional translation vector $\tau$; we can write
\begin{equation}
W = \begin{bmatrix}\tilde{S}&1\end{bmatrix}\begin{bmatrix}\tilde{M} & \tau\end{bmatrix}^T = SM^T,
\end{equation}
where $W$ now has rank at most $4$ and the constant ones vector is necessarily part of its column space.
We use this fact to naturally deal with offset data in our algorithms (Section~\ref{sec:all-ones}).

In the presence of noise, 
one can use the singular value decomposition (SVD)~\cite{tomasi1992shape}
in order to find $S$ and $M$
such that $SM^T$ most closely approximates $W$ in either the Frobenius or operator norm.  
In this paper we address the problem of
online structure from motion in which we have an estimated
factorization at time $t$ --- namely, $\hat{W}_t = \hat{S}_t\hat{M}^T_t$ --- and
we wish to update our estimate at time $t+1$ as we track points into the
next video frame. The new information takes the form of two additional
columns for $W_t$, leading to successive updates of the form  $W_{t+1} = \begin{bmatrix} W_t & v_t \end{bmatrix}$.
%\begin{equation}
%W_{t+1} = \begin{bmatrix} W_t & v_t \end{bmatrix}.
%\end{equation}
The algorithm for updating the approximate factorization must be
efficient enough to be used in real time, able to handle missing data
as points become occluded, and able to incorporate new points.

\section{Matrix Completion Algorithms for SFM}
\subsection{Background on the GROUSE Algorithm}
\label{sec:grouse}
We start by reviewing the GROUSE algorithm and its application to
structure from motion, in order to set the stage for our proposed
algorithm SAGE and its variants, which are discussed in later
subsections.

As in Section~\ref{sec:sfmdescription}, we use the subscript $t$ to
denote values at time $t$. Let $n_t$ be the number of tracks started
up to time $t$, let $U_t \in \mathbb{R}^{n_t\times 4}$ be a matrix 
with orthonormal columns that span the rank-4 column space of the
measurement matrix $W_t$, and let $v_t$ be a new column such that
$W_{t+1} = \begin{bmatrix}W_t &v_t\end{bmatrix}$. (We consider the two
  new columns provided by each video frame one at a time.) The set of
  observed indices of $v_t$ is denoted by $\Omega_t \subseteq
  \{1,\dots,n_t\}$, and $v_{\Omega_t}$ and $U_{\Omega_t}$ are row
  submatrices of $v_t$ and $U_t$ that correspond to $\Omega_t$.

The GROUSE algorithm~\cite{balzano_grouse, grouseconvergence} measures
the error between the current subspace-spanning matrix $U_t$ and the
new vector $v_t$ using the squared-$\ell_2$ distance
\begin{equation}
\label{eq:error}
{\mathcal E}(U_{\Omega_t},v_{\Omega_t}) = \|U_{\Omega_t} w_t - v_{\Omega_t}\|_2^2,
\end{equation}
where\footnote{Our notation $A^+b$ denotes the least-residual-norm solution to the
  least-squares problem $\min_x \, \| Ax-b \|_2^2$, obtainable from a
  singular value decomposition of $A$ or a factorization of $A^TA$.}
\begin{equation}
\label{eq:w}
w_t = U_{\Omega_t}^+v_{\Omega_t}
\end{equation} 
is the set of weights that project $v_{\Omega_t}$ orthogonally onto
the subspace given by $U_{\Omega_t}$.  We would like to replace $U_t$
by an updated matrix with orthonormal columns of the same dimensions, that reduces
the error ${\mathcal E}$ in accordance with the new observation $v_t$.
Denoting by $r_t$ the residual vector for the latest observation, we
define the subvector corresponding to the indices in $\Omega_t$ by
\begin{equation}
\label{eq:r}
r_{\Omega_t} = v_{\Omega_t} - U_{\Omega_t}w_t,
\end{equation}
and set the components of $r_t$ whose indices are not in $\Omega_t$ to
zero.
%$r_{\Omega_t^C} = 0$
%otherwise,
%\begin{equation}
%\label{eq:r}
%r_t(i) = 
%\begin{cases}
%v_{\Omega_t} - U_{\Omega_t}w_t & \text{if } i \in \Omega_t\\
%0 & \text{otherwise},
%\end{cases}
%\end{equation}
We can express the sensitivity of the error ${\mathcal E}$ to $U_t$ as
follows:
\begin{equation} \label{eq:deriv.E}
\frac{\partial \mathcal{E}}{\partial U_t} = -2r_tw_t^T,
\end{equation}
GROUSE essentially performs projected gradient descent on the
Grassmann manifold, taking a step in the negative gradient direction 
along the Grassmann manifold to maintain
orthonormality; see~\cite{balzano_grouse} for details.
%The online update equation is:
%\begin{align} 
%U_{t+1} =  U_t + & \left((\cos(\eta_t \sigma)-1)U_t\frac{w_t}{\|w_t\|} \right. \nonumber \\ 
%&\left. - \sin(\eta_t \sigma) \frac{r_t}{\|r_t\|} \right) \frac{w_t^T}{\|w_t\|} \;. \label{eq:grouseupdate}
%\end{align}

In structure from motion it is known \emph{a priori} that $\text{rank}(W)\leq 4$, and
so a subspace estimation algorithm like GROUSE, which requires the rank (or an upper bound)
as an input parameter, seems applicable. However, 
%
%{Because it is known \emph{a priori} that $\text{rank}(W)\leq 4$,
%GROUSE can be applied directly to the factorization problem, but 
there are several issues that prevent GROUSE from being easily used
for online structure from motion. First, GROUSE only maintains an
estimate of the column space of $W_t$, so the corresponding matrix
$R_t \in \mathbb{R}^{2m_t\times 4}$ for which $\hat{W_t} = U_tR_t^T$
must be computed whenever a final reconstruction is needed.  This can
be a problem for online applications since it requires keeping all
data until the algorithm is complete.  Additionally, the choice of a
critical parameter in GROUSE --- the step size for gradient descent
--- can affect the rate of convergence strongly.  Both issues
  would be resolved in an algorithm based on matrix factorization, but
  standard algorithms of this class do not handle missing or streaming
  data. Our SAGE algorithm, discussed below, leverages the
  relationship of GROUSE and incremental SVD to resolve these
  difficulties.
%% , and in Section~\ref{sec:sage} we take advantage of
%%   this approach to develop the algorithm SAGE. }

%\laura{Finally, we have seen empirically that GROUSE and other matrix
%completion algorithms have convergence rates sensitive to the spread
%in true singular values. This is a problem for SFM, as ... }
%In the following section, we take advantage of a recent result on the
%relationship between GROUSE and incremental SVD  to resolve
%these difficulties. The resulting algorithm is our base algorithm, SAGE.

\subsection{The Missing Data Incremental SVD (MD-ISVD) Formulation}
\label{sec:isvd}

The incremental SVD algorithm \cite{Bunch78} is a simple method for
computing the SVD of a collection of data by updating an initial
decomposition one column at a time. Given a matrix $W_t$ of rank $k$ whose thin SVD is $W_t
= U_t \Sigma_t V_t^T$, we wish to compute the SVD of a new matrix with
a single column added: $W_{t+1} = \begin{bmatrix} W_t &
  v_t\end{bmatrix}$.  Defining $w_t=U_t^T v_t$ and $r_t = v_t -
  U_tw_t$, we have
\begin{equation}
W_{t+1} = \begin{bmatrix} U_t & \frac{r_t}{\|r_t\|}\end{bmatrix} \begin{bmatrix} \Sigma_t & w_t \\ 0 & \|r_t\|
\end{bmatrix} \begin{bmatrix} V_t^T & 0 \\ 0 & 1 \end{bmatrix}\;,
\end{equation} 
where one can verify that the left and right matrices still have
orthonormal columns. An SVD of the center matrix yields
\begin{equation} \label{eq:isvdsigma}
\begin{bmatrix} \Sigma_t & w_t \\ 0 & \|r_t\|\end{bmatrix} = \tilde{U}\tilde{\Sigma}\tilde{V}^T,
\end{equation} 
where $\tilde{U}$, $\tilde{\Sigma}$, and $\tilde{V}$ are of size $(k+1) \times (k+1)$.

Two changes to ISVD are needed to make it suitable for online matrix
completion. First, to handle missing data,
% when only entries $\Omega_t \subseteq \{1,2,\dotsc,n_t\}$ are observed, 
we define the weights and residual vector in these update formulae as
in Equations \eqref{eq:w} and \eqref{eq:r}, respectively.  Second,
ensure that $w_t$ and $r_{\Omega_t}$ are well defined by \eqref{eq:w}
and \eqref{eq:r}, we limit the rank to $k$ and compute the {\em thin
  SVD}~\cite{brand2006fast}. Since only the first $k$ singular vectors
are needed, the smallest singular value and its associated singular
vectors can be dropped. Let $\hat{U}$, $\hat{\Sigma}$, and $\hat{V}$
be the $(k+1)\times k$, $k\times k$ and $(k+1)\times k$ matrices
resulting from this process. The updated rank-$k$ SVD estimate is then
given by $W_{t+1} = U_{t+1} \Sigma_{t+1} V_{t+1}^T$, where
%\vspace{-4mm}
\begin{align}
U_{t+1} = \begin{bmatrix} U_t  & \frac{r_t}{\|r_t\|}\end{bmatrix} \hat{U} \quad ; \quad \Sigma_{t+1}=\hat{\Sigma} \quad ; \quad V_{t+1} = \begin{bmatrix} V_t & 0 \\ 0 & 1 \end{bmatrix} \hat{V}\;. \label{eq:isvdupdate}
\end{align}

We call the resulting algorithm MD-ISVD (for ``missing-data
  ISVD''), and note that this algorithm has been previously proposed
  in \cite{brand2002incremental,grouseisvd}. Although this algorithm
  improves over ISVD by allowing missing data, it has been observed
  empirically~\cite{grouseisvd} that the convergence of MD-ISVD is
  slow when even a modest fraction of the data is missing. The
  column-space update involves an estimate of the singular values
  $\Sigma_t$, and for larger singular values, the update will favor
  the corresponding columns of $U_t$, even though in early iterations
  those columns are poor estimates based on partial
  data. In~\cite{brand2002incremental}, this issue is addressed by
  down-weighting the singular values with a forgetting factor
  $\beta<1$. SAGE instead addresses this issue by being agnostic to
  singular values and using an identity matrix in the update, as we
  show next.

\subsection{The SAGE Algorithm}
\label{sec:sage}

We now examine the relationship of this MD-ISVD update to the GROUSE
update, in the context of SFM. Let $\hat{W}_t = U_tR_t^T$ be an
estimated rank-4 factorization of $W_t$ such that $U_t$ has
orthonormal columns.  Given a new column $v_t$ with observed entries
$\Omega_t$,
% let $W_{t+1}
%= \begin{bmatrix} \hat{W}_t & v_t\end{bmatrix}$ be the updated matrix.
  if $w_t$ and $r_t$ are the least-squares weight and residual vector,
  respectively, defined
with respect to the set of observed indices $\Omega_t$ as in Equations
\eqref{eq:w} and \eqref{eq:r},
%If
%$w_t = U_{\Omega_t}^+v_{\Omega_t}$ is the set of 
%weights that give the orthogonal projection of $v_{\Omega_t}$ onto $U_{\Omega_t}$, and $r_{\Omega_t} = v_{\Omega_t} - U_{\Omega_t}w_t$ is the
%resulting residual vector on $\Omega_t$ with $r_{\Omega_t^C}=0$, 
then we can write
\begin{equation}
\label{eq:grouseisvd}
\begin{bmatrix}U_t R_t^T & \tilde{v}_t \end{bmatrix} = \begin{bmatrix}U_t & \frac{r_t}{\|r_t\|}\end{bmatrix}
 \begin{bmatrix}I & w_t\\ 0 & \|r_t\|\end{bmatrix}\begin{bmatrix}R_t &0\\0&1\end{bmatrix}^T,
\end{equation} 
% which can be easily verified by multiplying the matrices on the right-hand side. 
where the subvector of $\tilde{v}_t \in \mathbb{R}^{n_t}$
corresponding to $\Omega_t$ is set to $v_{\Omega_t}$ while the
remaining entries are imputed as the inner product of $w_t$ and the
rows $i \notin \Omega_t$ of $U_t$. Stated formally, letting
$[\tilde{v}_t]_i$ refer to the $i^{th}$ component of the vector
$\tilde{v}_t$, and we have:

$$[\tilde{v}_t]_i := \left\{ \begin{matrix} [v_t]_i & i \in \Omega_t \\ [U_t w_t]_i  & i \in \Omega_t^C \end{matrix} \right.\;.$$
%
%
%In the case of missing data, $w_t$ and $r_t$ can again be defined
%with respect to the set of observed indices $\Omega_t$ as in Equations
%\eqref{eq:w} and \eqref{eq:r}. 
%
Define the SVD of the center matrix in Equation~\eqref{eq:grouseisvd}
to be
\begin{equation} \label{eq:identitysvd}
\begin{bmatrix}I & w_t\\0 & \|r_t\|\end{bmatrix} = \tilde{U}\tilde{\Sigma}\tilde{V}^T.
\end{equation}
Let $\hat{U}$ and $\hat{V}$ be the $5\times 4$ matrices obtained by
dropping the last columns of $\tilde{U}$ and $\tilde{V}$,
respectively, corresponding to the smallest singular value of
$\tilde{\Sigma}$. Similarly, let $\hat{\Sigma}$ be the $4\times 4$
diagonal matrix obtained by dropping the last column and row from
$\tilde{\Sigma}$.

In \cite{grouseisvd}, it was shown that updating $U_t$ to
\begin{equation} \label{eq:isvdUupdate}
U_{t+1} = \begin{bmatrix} U_t & \frac{r_t}{\|r_t\|}\end{bmatrix}\hat{U}  
\end{equation} 
(similarly to the first equation in Equation~\eqref{eq:isvdupdate}) is
equivalent to GROUSE for a particular data-dependent choice of step
size for the gradient algorithm. We use this update as the
  basis for SAGE.

Additionally, combining Equations~\eqref{eq:grouseisvd} and~\eqref{eq:identitysvd},
%because
%\vspace{-4mm}
%\begin{equation}
%\begin{bmatrix}U_tR_t^T & v_t \end{bmatrix} =  \begin{bmatrix}U_t & \frac{r_t}{\|r_t\|}\end{bmatrix} \tilde{U}\tilde{\Sigma}\tilde{V}^T
%\begin{bmatrix}R_t &0\\0&1\end{bmatrix}^T,
%\end{equation}
we may update $R_t$ as follows:
\begin{equation} \label{eq:isvdRupdate}
R_{t+1} = \begin{bmatrix}R_t & 0 \\ 0 & 1\end{bmatrix} \hat{V} \hat{\Sigma}.
\end{equation}
The result is a new rank-4 factorization $\hat{W}_{t+1} =
U_{t+1}R_{t+1}^T$. Both $U_{t}$ and $R_{t}$ can be maintained with
every update, so the estimated $\hat{W}_t$ is available in real-time.
With further modifications to handle the addition of rows to the
matrix and the all-ones offset vector in the column space (see
Section~\ref{sec:implementation}), we may apply this algorithm to
real-time structure from motion.  We call this algorithm ``SAGE.''
%and it is described again below, in Algorithm~\ref{alg:meta} and Table~\ref{tbl:family}.

%The advantages of this method are two-fold.  
We now comment on the advantages of SAGE. First, by updating
  both $U$ and $R$ simultaneously, there is no need to calculate $R$
  whenever reconstruction is needed. Instead, we keep a running
  estimate of both $U$ and $R$ so estimates of the motion and structure
  matrices can easily be produced at any point in time. While the ISVD
  algorithm also updates both matrices, it can not handle missing data,
  and while the GROUSE algorithm handles missing data, it can not
  update both $U$ and $R$ simultaneously.  With SAGE we thus have a
  truly online factorization-based SFM approach, because we no longer need to store the
  entire observation matrix $W$ in order to solve for $R$. Keeping a
subset of the data may still be useful, so that old data can be
revisited and reused when there is time available in the
computation. The amount of data stored for this purpose may be limited
to a fixed amount to constrain the memory footprint. Second, this formulation
uses an implicit step size, as it is a matrix factorization approach
as opposed to a gradient descent approach.  Therefore we no longer are
required to specify a step size as in the original GROUSE formulation,
although the residual vector can still be scaled to change the
effective step size if needed. 
%In Section~\ref{sec:experiments}, we
%show that SAGE outperforms all other batch and online algorithms for
%SFM on both real and synthetic data in terms of both error and speed.

%\footnote{SJW: Which algorithm are you talking about here? GROUSE or ISVD or both?}
% An explicit value for
% this step size is given in \cite{grouseconvergence}, and we found that
% it works well in practice.
%\laura{Finally, this perspective leaves open the possibility 
%to also track the singular value matrix $\Sigma$, which allows us potentially more flexibility with poorly conditioned measurement matrices.}

\subsection{RSAGE for Robust SFM}
\label{sec:robust_grasta}
% We can  use grasta in there as well to deal with outliers
The GRASTA algorithm~\cite{he2012cvpr,he2011grasta} is an extension of GROUSE that is robust to outliers. 
Instead of minimizing the $\ell_2$ cost function given in Equation \eqref{eq:error}, 
GRASTA uses the robust $\ell_1$ cost function
$${\mathcal E}_{grasta}(U_{\Omega_t},v_{\Omega_t}) = \min_{w_t} \|U_{\Omega_t} w_t - v_{\Omega_t}\|_1\;.$$ 
%In order to minimize this, 
%Grasta uses the augmented Lagrangian form of the cost function,
%\begin{align}
%	\mathcal{L} (U_t, s_t, w_t, y_t) = {\| s_t \|}_1 &+& y_t^T(U_{\Omega_t} w_t + s_t - v_{\Omega_t})\nonumber\\ &+& \frac{\rho}{2}\| U_{\Omega_t} w_t + s_t - v_{\Omega_t} \|_2^2 \label{eq:L1_aug_Lagrangian}
%\end{align}
%where $s_t \in \R^{|\Omega_t|}$ is the sparse additive outlier component and $y_t\in \R^{|\Omega_t|}$ is the dual vector. 
GRASTA estimates the weights $w_t$ of this $\ell_1$ projection as well
as the sparse additive component using ADMM~\cite{boyd2010distributed},
 then updates $U_t$ using Grassmannian geodesics, replacing $r_t$
with a variable $\Gamma_t$, which is a function of the sparse
component and the $\ell_1$ weights~\cite{he2012cvpr}.
%\footnote{SJW: I wonder if you could use a lower-case Roman character here, rather than an upper-case Greek, more consistent with the vector notation used elsewhere in the paper.} 
We can use these $w_t$ and $\Gamma_t$ in place of $w_t$ and
  $r_t$ in SAGE (Section~\ref{sec:isvd}), resulting in a robust
  variant we call RSAGE. The novelty in RSAGE is to incorporate these
  $\ell_1$ weights and residuals into a matrix factorization
  framework. In Section \ref{sec:experiments}, we show that RSAGE is better able to find
the correct 3D structure in the presence of sparse outliers. 
While outside the scope of this paper, we believe this
  merging of ADMM-computed weights and residuals with a matrix
  factorization algorithm is worthy of more extensive study.

\subsection{SAGE, RSAGE, and MD-ISVD: A Family of Algorithms}
\label{sec:family}

The algorithms that we have presented so far make up a family of
algorithms that can be applied to SFM in various circumstances. We
study SAGE and RSAGE along with MD-ISVD, which carries forward an
estimate for the singular values, $\Sigma$.

Another variant within this family could be obtained by scaling the
residual norm in~\eqref{eq:identitysvd} by some value $\alpha_t$
before taking its SVD:
\begin{equation} \label{eq:identityscaled}
\begin{bmatrix}I & w_t\\ 0 & \alpha_t \|r_t\|\end{bmatrix} = \hat{U}\hat{\Sigma}\hat{V}^T\;.
\end{equation}
This scaling potentially yields more control over the contribution of
the residual vector to the new subspace estimate. We tried decreasing
the value $\alpha_t$ with increasing $t$, and found this approach to
be useful the batch setting. However, for online data, the no-scaling
choice ($\alpha_t=1$) performs best.
%\footnote{SJW: Could you clarify here that ``no scaling'' means $\alpha_t \equiv 1$?}

After we describe another relevant algorithmic component in the next section, the incorporation of
an offset vector into the column space,
we present in Algorithm~\ref{alg:meta} one iteration of each algorithm
in the family discussed in this section. Table~\ref{tbl:family} shows
the specifics of all proposed algorithms relative to this
meta-algorithm.

%\vspace{-5pt}
\section{Implementation} \label{sec:implementation}
\subsection{The All-1s Vector}
\label{sec:all-ones}

A few more issues need to be addressed for implementation of SAGE for
SFM. The first issue is to exploit the fact that the constant vector
of ones (denoted as $\mathds{1}$) should always be in the column space of $W_t$ and thus should
be in the span of any estimate $U_t$ of the column space of $W_t$,
since the points may be offset from the origin.  Without loss of
generality, suppose that $U_t$ has the ones-vector as its last column
(appropriately scaled), so that
\begin{align}
U_t = \begin{bmatrix}\bar{U}_t & \mathds{1}/\sqrt{n_t}\end{bmatrix}, \quad R_t = \begin{bmatrix}\bar{R}_t &\tau_t\sqrt{n_t}\end{bmatrix}.
\end{align}
Here $\tau_t$ is the corresponding translation vector, which is the
(scaled) last column of $R_t$.  Similarly, let $w_t = \begin{bmatrix}
  \bar{w_t}^T&\gamma_t\end{bmatrix}^T$.  The derivative of the error
  $\mathcal{E}$ in Equation \eqref{eq:error} with respect to just the first
  three columns is 
\begin{equation}
\frac{\partial \mathcal{E}}{\partial \bar{U_t}} = -2r_t\bar{w_t}^T
\end{equation}
(cf. \eqref{eq:deriv.E}). By not considering the derivative of $\mathcal{E}$ with respect to the
ones vector it will remain in the span of $U$, since the SAGE update
will be applied only to $\tilde{U}$.
%We showed in section \ref{sec:grouse} that by using the residual
% vector $r$ and the truncated weight vector $\tilde{w}$, performing a
% grouse update only on $\tilde{U}$ ensures that the constant ones
% vector remains in the span of $U$.
Our SAGE update for structure from motion is obtained by first setting
\begin{align}
\bar{U}_{t+1} = \begin{bmatrix} \bar{U}_t & \frac{r_t}{\|r_t\|}\end{bmatrix}\tilde{U}\quad;\quad
\bar{R}_{t+1} = \begin{bmatrix}\bar{R}_t & 0 \\ 0 & 1\end{bmatrix} \tilde{V} \tilde{\Sigma}\quad;\quad
\tau_{t+1} = \begin{bmatrix}\tau_t\\\gamma_t/\sqrt{n_t}\end{bmatrix}\;,
\end{align}
%\footnote{SJW: Could a subequations format be used here, so these three equations are numbered a,b,c? They are not labelled - do they even need to be numbered/ Perhaps put them all on one unnumbered line, or use environment align* instead?}
(where $\tilde{U}$, $\tilde{\Sigma}$, and $\tilde{V}$ are defined as in
Equation \eqref{eq:identitysvd} using $\bar{w}_t$), and then dropping the last column
of $\bar{U}_{t+1}$ and $\bar{R}_{t+1}$.
%% \begin{equation}
%% \begin{bmatrix}I & \tilde{w}_t\\0 & \|r_t\|\end{bmatrix} = \hat{U}\hat{\Sigma}\hat{V}^T.
%% \end{equation}
Because the residual vector $r_t$ is still based on the full matrix
$U_t$, including the ones-vector, $r_t$ will necessarily be orthogonal
to the ones-vector. Therefore, since $$U_{t+1} =  \begin{bmatrix}\bar{U}_{t+1} & \mathds{1}/\sqrt{n_{t}}\end{bmatrix}\;,$$ 
$U_{t+1}$ will retain the ones-vector in its
span and will still have orthonormal columns.
%Thus, by using the full residual vector $r_t$ and the truncated
%weight vector $\tilde{w}$, the ones vector will be implicitly
%maintained as an additional column.

The same process can be applied to MD-ISVD. However, we note that the
resulting method is no longer a true SVD, in that the updated
$R_{t+1}$ will not necessarily have orthonormal columns. (In
particular, the last column $\tau_t$ may not be orthogonal to the
other columns, which make up the submatrix $\bar{R}_{t+1}$.)
%%  in the following sense. In
%% the standard ISVD algorithm, both $U_t$ and $R_t$ are always
%% orthogonal matrices.  When the all-$1$'s vector is constrained to be
%% the last column of $U_t$, however, then the SVD update only ensures
%% that the submatrix $\bar{R}_t$ is an orthogonal matrix. That is,
%% $\tau_t$ (the last column of $R_t$) may no longer be orthogonal to the
%% remaining columns of $R_t$, which form $\bar{R}_t$. 
In our experiments, all algorithms are constrained to maintain the
all-$1$'s vector in their column space.  Although this is no longer a
true SVD, we found that it leads to better solutions than simply
finding a rank-$4$ approximation. This observation has been reported
previously, for example, in \cite{buchanan2005damped}, where the best
rank-$4$ factorization of the Dinosaur dataset was found to produce a
suboptimal 3D model.

\begin{algorithm}[t]
\caption{One Iteration of the Meta-Algorithm} \label{alg:meta}
\begin{algorithmic}
    \STATE{{\bf INPUT}:
        \begin{itemize}[itemsep=0pt, topsep=0pt, partopsep=0pt]
\item $U_t,D_t,$ and $R_t$ of the appropriate algorithm (as given by Table~\ref{tbl:family}) for the current matrix estimate $\hat{W}_t = U_tD_tR_t^T$, where
\[ 
U_t = \begin{bmatrix}\bar{U}_t & \mathds{1}/\sqrt{n_t}\end{bmatrix}\quad ; \quad D_t = \begin{bmatrix}\bar{D}_t & 0\\0 &  1 \end{bmatrix} \quad ; \quad  R_t = \begin{bmatrix}\bar{R}_t & \tau_t\sqrt{n_t}\end{bmatrix}; 
\]
\item scaling parameter $\alpha_t$;\vspace{-5pt}
\item the new column $v_{\Omega_t}$ with only points indexed by $\Omega_t$ observed. 
\end{itemize}
}
\STATE{Calculate the weights $w_t = \begin{bmatrix} \bar{w}_t & \gamma_t\end{bmatrix}^T$ and residual $r_t$ for the appropriate algorithm according to Table~\ref{tbl:family}\;; }
\STATE{ Define $\bar{v}_t = \bar{U}_t\bar{w}_t + r_t$ to be the portion of the imputed 
vector $v_t = U_tw_t + r_t$ orthogonal to $\mathds{1}$\;; }
\STATE{Noting that
\begin{equation}
\begin{bmatrix} \bar{U}_t\bar{D}_t\bar{R}_t^T & \bar{v}_t\end{bmatrix}
 = 
\begin{bmatrix} \bar{U}_t & \frac{r_t}{\|r_t\|}\end{bmatrix}
\begin{bmatrix} \bar{D}_t & \bar{w}_t \\ 0 & \|r_t\|\end{bmatrix}
\begin{bmatrix} \bar{R}_t & 0 \\ 0 & 1 \end{bmatrix}^T,
\end{equation}
compute the SVD of the update matrix (with scaled residual):
%\footnote{SJW: I removed ``with singular values in decreasing order'' since that it part of the definition of SVD.}
\begin{equation}
\begin{bmatrix}\bar{D}_t & \bar{w}_t \\ 0 & \alpha_t\|r_t\| \end{bmatrix} = 
\tilde{U}_t\tilde{\Sigma}_t\tilde{V}_t^T\;;
\end{equation}
}
\STATE{ Update the column space and translation vector:
\begin{align}
\bar{U}_{t+1} = \begin{bmatrix} \bar{U}_t & \frac{r_t}{\|r_t\|}\end{bmatrix}\tilde{U}_t
\quad ; \quad
{\tau}_{t+1} = \begin{bmatrix} \tau_t \\ \gamma_t/\sqrt{n_t} \end{bmatrix}\;;
\end{align}
}
\STATE{Update $\bar{R}_{t+1}$ and $\bar{D}_{t+1}$ for the appropriate algorithm according to Table~\ref{tbl:family}\;; }
\STATE{Drop the last column of both $\bar{U}_{t+1}$ and $\bar{R}_{t+1}$. Drop the last
column and the last row of $\bar{D}_{t+1}$\;;}
\STATE{Update
\begin{align}
U_{t+1} = \begin{bmatrix} \bar{U}_{t+1} & \mathds{1}/\sqrt{n_{t}}\end{bmatrix} 
\quad ; \quad
D_{t+1} = \begin{bmatrix} \bar{D}_{t+1} & 0 \\ 0 & 1 \end{bmatrix}
\quad ; \quad
R_{t+1} = \begin{bmatrix} \bar{R}_{t+1} & \tau_{t+1}\sqrt{n_t}\end{bmatrix}\;;
\end{align}
}
\STATE{{\bf OUTPUT}: $\hat{W}_{t+1} = U_{t+1}D_{t+1}R_{t+1}^T$.}
\end{algorithmic}
\end{algorithm}
%\footnote{SJW: The ``align'' environment is used a lot in this
%  algorithm but no labels are given so the equations can not be
%  referenced. Replace these by the unnumbered display equation
%  environment?}

\begin{table}[h]
    \footnotesize
  \centering
  \begin{tabular}{c|c|c|c}
	\toprule
	\multicolumn{4}{c}{{\bf A Family of  Matrix Completion SFM Algorithms}} \\
	\midrule 
	& {\bf SAGE ($\ell_2$)} & {\bf RSAGE ($\ell_1$)} & {\bf MD-ISVD}  \\
	\midrule 
	\multicolumn{4}{l}{{\bf Matrix information}}\\	
	\midrule
	$\bar{D}_t=$ & $I_{k-1}$ & same as SAGE & $\Sigma_t$, an estimate of singular values\\
	\midrule
	$\bar{R}_t$ orthogonal? & no & no & yes \\
	\midrule 
	\multicolumn{4}{l}{{\bf For Algorithm \ref{alg:meta}}}\\
	\midrule
	$w_t,r_t$ & $w_t=\arg\min_w \|U_{\Omega_t}w - v_{\Omega_t}\|_2^2 $& computed using &  either SAGE or RSAGE\\
		  & $r_{\Omega_t} = v_{\Omega_t}-U_{\Omega_t}w_t\;;\;r_{\Omega_t^C} = 0$ & ADMM (Section \ref{sec:robust_grasta}) &depending on cost\\
	\midrule
	Update $\bar{R}_{t+1}$ & $\bar{R}_{t+1} = \begin{bmatrix}\bar{R}_t & 0 \\ 0 & 1\end{bmatrix}\tilde{V}_t\tilde{\Sigma}_t$ & same as SAGE & $\bar{R}_{t+1} = \begin{bmatrix}\bar{R}_t & 0 \\ 0 & 1\end{bmatrix}\tilde{V}_t$ \\
	\midrule
	Update $\bar{D}_{t+1}$ & $\bar{D}_{t+1} = I_{k-1}$ & same as SAGE &  $\bar{D}_{t+1} = \tilde{\Sigma}_t$ \\
	\midrule 
	%Update $\tau_{t+1}$ & $\tau_{t+1} = \begin{bmatrix}\tau_t\\ \gamma_t\end{bmatrix}$& same as GROUSE & $\tau_{t+1} = \frac{1}{\sqrt{1+\gamma_t^2}}\begin{bmatrix}\tau_t \\ \gamma_t\end{bmatrix}$\\
	%\midrule
	%Update $\delta_{t+1}$ & $\delta_{t+1} = 1$ & same as GROUSE & $\delta_{t+1} = \delta_t\sqrt{1+\gamma_t^2}$\\
	%\midrule
	\multicolumn{4}{l}{{\bf For Algorithm \ref{alg:outerloop}}}\\
	\midrule
	Initialization restrictions& $U_0=\begin{bmatrix}\bar{U}_0 & \mathds{1}/\sqrt{n_0}\end{bmatrix}$ orthogonal\; & same as SAGE & $U_0=\begin{bmatrix}\bar{U}_0 & \mathds{1}/\sqrt{n_0}\end{bmatrix}$ orthogonal\\
	   & $D_0 = I_{k-1}$ \; & & $D_0 = \begin{bmatrix}\bar{D}_0 & 0 \\ 0 & 1\end{bmatrix}$ positive, diagonal\\
                       & $R_0$ unrestricted & & $\bar{R}_0$ orthogonal\\
	\midrule
	Remove a row of $R_t$ & just remove it & same as SAGE & downdate as in \cite{brand2006fast}, then remove it\\
	\bottomrule 
  \end{tabular} 
  \caption{Algorithmic specifics in each of our family of matrix completion algorithms
    for structure from motion, as used in Algorithm~\ref{alg:meta} and~\ref{alg:outerloop}.}
  \label{tbl:family}
  \end{table}

\subsection{Adding New Points} \label{sec:addingpts}
% A second issue for implementation is the fact that 
In online structure
from motion, we are initially unaware of the total number of points
to be tracked and need to account for newly added points as the
video progresses. If $U_t \in \mathbb{R}^{n_t\times 4}$ is the current
subspace estimate, then new points will manifest themselves as
additional rows of $U_t$.  However, when updating $U_t$, we have to
make sure that the columns of $U_t$ remain orthonormal and the last
column continues to be the vector of ones.  We thus perform the following
update when each new point is added, where we increment $n_{t+1} = n_t + 1$ so that the columns of 
$U_t$ remain orthonormal:
\begin{equation} \label{eq:addingpts1}
U_t  \leftarrow \begin{bmatrix} \bar{U}_t& \mathds{1}/\sqrt{n_t+1} \\ 0& 1/\sqrt{n_t+1}\end{bmatrix}.
\end{equation}
We also update $\tau_t$, the last column of $R$, so that their product remains the same for current
entries:
\begin{equation} \label{eq:addingpts2}
\tau_t \leftarrow \tau_t \sqrt{\frac{n_t+1}{n_t}}\;.
\end{equation}
%By appending zeros to each column of $\bar{U_t}$, they remain
%orthonormal.  
%In the supplementary material we provide an alternative
%way to add new points, but do not use this method in our experiments.

\subsection{Updating Past Points}
The SAGE update for structure from motion is fast enough that many
updates can be done for each new video frame.  It is therefore
advantageous to be able to revisit old frames and reduce the error
more than would be possible using a single pass over the frames.
Using the original SAGE formulation described in Section~\ref{sec:grouse}, we simply run
additional SAGE updates using past columns of $W$.  This does not
work in the new online formulation, where we also keep track of the
matrix $R_t$, since running another SAGE update will add a new row
to $R_t$.  Instead, we simply drop the associated row of $R_t$ before
the update and replace it with the resulting new row.  Because we do
not impose any orthogonality restrictions on the matrix $R_t$, no correction is
needed. In our experiments, however, we also compare to MD-ISVD, which
requires that the the right-side matrix be orthogonal.  
 (We ``downdate'' our SVD using the algorithm given by
Brand~\cite{brand2006fast} before performing an update.)

Having discussed these implementation details, we can define the outer loop of our algorithm, given in Algorithm~\ref{alg:outerloop}. This outer loop calls Algorithm~\ref{alg:meta} to perform each iteration. %This outer loop is described in Algorithm~\ref{alg:outerloop}.

\begin{algorithm}[h]
\caption{The outer loop of our SFM algorithm} \label{alg:outerloop}
\begin{algorithmic}
\STATE{{\bf INPUT}:
\begin{itemize}
\item initial matrix $W_0$ of size $n_0\times 2m_0$;
\item a sequence of scaling parameters $\alpha_0, \alpha_1,\dotsc$;
\item desired rank $k$ ($k=4$ for SFM).
\end{itemize}
}
\STATE{Initialize $U_0$ (size $n_0 \times k$), $D_0$ (size $k\times k$), and $R_0$ (size $2m_0\times k$) for the appropriate algorithm according to Table~\ref{tbl:family}\;;}
\STATE{Set $t=0$, $T=0$\;;}
\REPEAT 
\STATE{Take the new column $v_{\Omega_t}$ with only points indexed by $\Omega_t$ observed\;;}
\IF{$\Omega_t$ includes $\kappa > 0$ previously unobserved rows}
\STATE{Update $U_t,\tau_t$ as in Equations~\eqref{eq:addingpts1} and \eqref{eq:addingpts2} and increment $n_{t} = n_{t-1} + \kappa$ ;}
\ELSE
\STATE{Let $n_t = n_{t-1}$ ;}
\ENDIF
\IF{$t<T$}
\STATE{Remove row $t$ of $R_t$ for the appropriate algorithm according to Table~\ref{tbl:family}\;;}
\ENDIF
\STATE{Execute Algorithm~\ref{alg:meta} to update the matrix estimates ;}
\STATE{Move the last row of $R$ to row $t$\;;}
\IF{the next frame has yet to arrive for processing}
\STATE{Set $t$ to be a random integer from 1 to $T$ ;}
\ELSE
\STATE{Set $t=T+1$ and $T = T+1$ ;}
\ENDIF
\UNTIL{termination}
\end{algorithmic}
\end{algorithm}

\subsection{Complexity} \label{sec:complexity}

Finally we comment on the per-iteration computational complexity of SAGE, RSAGE, and MD-ISVD. Recall that at iteration $t$, $n_t$ is the number of features, $T$ is the number of frames so far, $|\Omega_t|$ is the number of observations in the frame being used for the update, and $k$ is the rank of the decomposition. The computation of the weights in Algorithm~\ref{alg:meta} is the only place where the three algorithms differ. For SAGE and MD-ISVD, we must solve a Least Squares problem for the weights, giving $O(|\Omega_t|k^2)$ operations. For RSAGE,  we use ADMM to compute the weights. Assuming that ADMM uses a constant number of iterations, this computation requires $O(|\Omega_t|k^3)$ operations. Computing the residual then requires $O(n_tk+|\Omega_t|)$ operations. To compute the SVD of the center $k \times k$ matrix we need $O(k^3)$ operations, and then updating $U$ and $R$ requires $O(n_tk^2)$ and $O(Tk^2)$ operations respectively. In the typical situation where $k < |\Omega_t| \ll n_t$, these final two computations are the most burdensome, giving an overall computational complexity of $O((n_t+T)k^2)$ for the $t^{th}$ iteration. 

We emphasize that this is only the {\em per iteration} complexity. For these algorithms, we have yet to analyze how many iterations are required for convergence. While SAGE and MD-ISVD have the same per-iteration complexity, they converge at very different speeds. Additionally, typical batch algorithms handle the whole matrix in every iteration, so the per-iteration complexity of SAGE is not directly comparable to the complexity given, \eg in~\cite{chen2008optimization}. Therefore, to show the practical impact on realtime structure from motion, we focus on runtime in our experiments.

\section{Experiments} \label{sec:experiments}
In this section we evaluate SAGE for rigid structure from motion in both the online and batch settings.

\subsection{Batch experiments}
\label{sec:batch}
\subsubsection{Algorithms}
Each of the algorithms we compare to was modified 
such that the constant ones-vector always remains in the column space of the estimated
matrix. The algorithms we compare are:
\begin{itemize}
\item {\bf SAGE} (SAGE and SAGE100) \cite{balzano_grouse}: The main algorithm presented in this paper. In the batch
setting, one iteration is defined as one pass over all columns of the measurement matrix in
a random order. In SAGE, the residuals are not manually scaled. In SAGE100, we add an additional
scaling of the residual in order to ensure
convergence. The residual was scaled by $\alpha_t = C/(C+t)$ at iteration $t$, where $C=100$. 
\item {\bf Power Factorization} (PF) \cite{hartley2003powerfactorization}: Alternating least-squares optimization of $U$ and $R$.
\item {\bf Guerreiro and Aguiar} (GA) \cite{guerreiro20023d}: Iteratively fill in the missing matrix values using
the current matrix estimate, and then find the best low-rank matrix with respect to this filled-in matrix.
\item {\bf Damped Newton} (DN) \cite{buchanan2005damped}: A second-order damped Newton's method with respect to all values of $U$ and $R$.
\item {\bf Levenberg-Marquardt Subspace} (LM\_S) \cite{chen2008optimization}: The cost function is redefined with
respect to only $R$, and Levenberg-Marquardt minimization is used.
\item {\bf Levenberg-Marquardt Manifold} (LM\_M) \cite{chen2008optimization}: Similar to LM\_S, but optimization
is restricted to remain on the Grassmann manifold.
\item {\bf Wiberg} (WIBERG) \cite{okatani2007wiberg}: A second-order method that linearizes the cost function
around $U$ and performs a Newton-type minimization with respect to $R$.
\item {\bf Damped Wiberg} (DW) \cite{okatani2011efficient}: A related method to the previous, this implementation adds a damping factor that allows for faster runtimes.
\item {\bf Missing Data Incremental SVD} (MD-ISVD) \cite{brand2002incremental, grouseisvd}: As presented in this paper, it is similar to SAGE but additionally keeps track of the singular values. At each iteration, the current column's data is
``downdated'' before being subsequently updated.
\item {\bf Column Space Fitting} (CSF) \cite{gotardo2011computing}: A Levenberg-Marquardt optimization method that encourages the matrices of camera parameters
to vary smoothly by restricting them to be composed of a basis of vectors in the discrete cosine transform domain.
Based on the default parameters used in the code and the notes in the paper, we began with a set of basis vectors
that is 10\% of the total number of frames, and the basis is enlarged by this same amount every 100 iterations until
the maximum size is reached.
\item{\bf Bilinear Modeling via Augmented Lagrange Multipliers} (BALM) \cite{del2012bilinear}: A model which restricts the camera parameter
matrices to lie on a particular manifold by projecting it onto the manifold after each iteration. Here, we use the scaled-orthographic projection. We update the Lagrange multipliers every 500 iterations. We found this number of iterations to perform
well on most datasets, but note that better convergence may be obtained by fine-tuning this parameter to each dataset.
For testing convergence, one iteration of BALM was defined as 50 internal iterations of the inner loop.
\end{itemize}

All algorithms were implemented within a common framework with a similar amount of optimization. All code is implemented using MATLAB, aside from one internal function for DN which is written in C++. The experiments were run on Amazon EC2 computers with 1.8GHz Intel Core i5 processors and 4Gb of RAM.

Note that CSF and BALM optimize somewhat different cost
  functions than the other algorithms, including SAGE. BALM has an
  additional constraint requiring the camera matrices to lie on a
  certain manifold. CSF uses a set of basis functions that assume the
  camera moves smoothly over time. All other algorithms attempt only
  to find a rank-4 factorization that minimizes the squared error of
  the observed entries, contingent on the constant ones vector being
  part of the factorization's column space. These algorithms do not
  impose constraints on the camera matrices themselves other than
  their rank, and thus they assume a general affine camera model. Only
  after a rank-4 solution is obtained do we impose scaled-orthographic
  metric constraints \cite{poelman1997paraperspective} to find a
  resulting 3D structure. We note that it is possible to incorporate
  additional constraints into SAGE. For example, just as is done in
  BALM, the solution could be projected onto an appropriate manifold
  after each iteration of SAGE, to ensure that the camera matrices are
  of the proper form. However, we did not encounter issues with
  degeneracy in our experiments, so did not find it necessary to
  implement this variant.
 %% and adding such a step would also add
 %%  to SAGE's computation time.}

We imposed the same constraint on all rank-4 factorizations to include
the all-ones vector. Note that several of the algorithms \cite{okatani2011efficient,buchanan2005damped,chen2008optimization} previously report results that 
instead use general rank-4 factorizations without imposing the all-ones vector and subsequently
measure only 2D RMSE.
While this did not greatly affect convergence time, 
we found that the inclusion of the all-ones vector constraint
may cause algorithms to get suck 
in local minima on some datasets (in terms of 2D RMSE). However, if the all-ones vector
is not enforced, it would have to be enforced after convergence in order to obtain the final 3D model.
A more in-depth comparison of the convegence of algorithms under different such constraints
is worthy of further research.

All algorithms were run until convergence from $100$ different random
initializations. For CSF, a ``random'' initialization is not as
straightforward because of its basis functions, and we initialized it
by using it to approximate the random matrix and then using the
resulting parameters as its initialization. We also evaluated each
method using a deterministic initialization where values were filled
in using the mean value of the corresponding column. For CSF, the
deterministic initialization consists of using the default
initialization as described in \cite{gotardo2011computing}.

We declared convergence to have occurred when the 2D RMSE did not
decrease more than $1\%$ in the previous $10$ minutes or $10$
iterations. For the synthetic dataset, we also measure 3D RMSE. These
3D reconstructions were created by applying standard
scaled-orthographic metric constraints
\cite{poelman1997paraperspective} and aligning them to the groundtruth
model using a Procrustes transformation. 

Note that while the datasets used here are captured with standard projective cameras, the factorization
model we use assumes an affine camera. However, we have chosen datasets that have minimal 
perspective distortions, some of which have been previously
used in factorization-based algorithms \cite{fitzgibbon1998automatic, tardif2007algorithms}. For videos that contain
significant perspective distortions, other methods might be more appropriate.

\begin{figure}[t]
\begin{center}
\hspace{-25pt}
\begin{subfigure}{4.6cm}
\includegraphics[width=4.6cm]{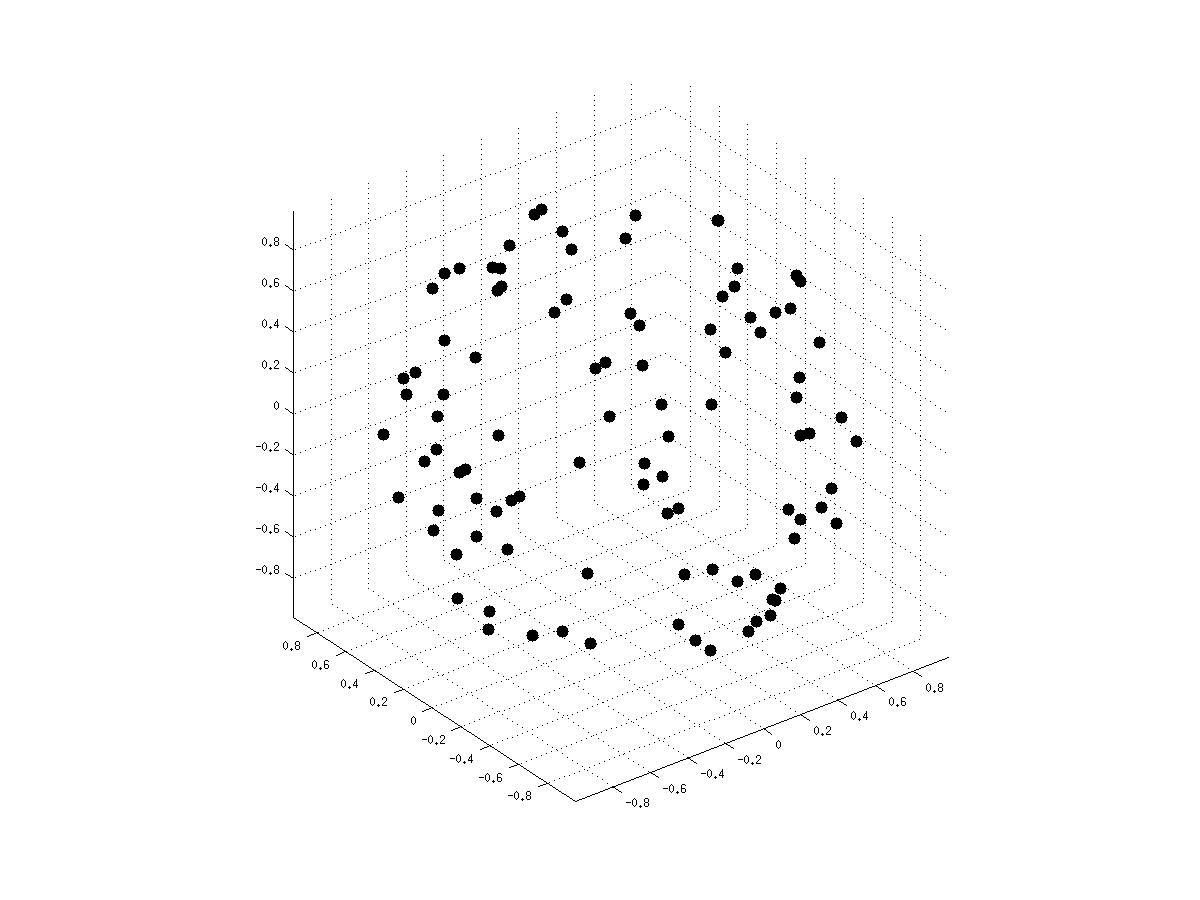}
\caption{}
\end{subfigure}\hspace{-20pt}
\begin{subfigure}{4.6cm}
\includegraphics[width=4.6cm]{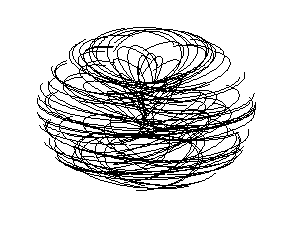}
\caption{}
\end{subfigure}
\hspace{-20pt}
\begin{subfigure}{4.6cm}
\includegraphics[width=4.6cm]{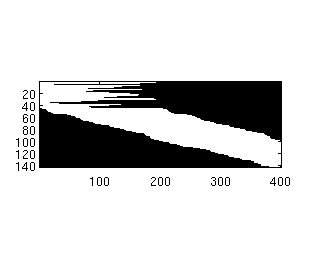}
\caption{}
\end{subfigure}
\hspace{-25pt}
\begin{subfigure}{4.6cm}
\includegraphics[width=4.6cm]{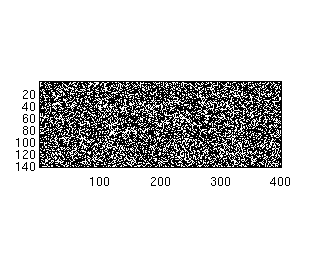}
\caption{}
\end{subfigure}
\end{center}
\caption{{\bf Synthetic Sphere dataset}. {\bf (a)} 3D model of the sphere. {\bf (b)} Observed point tracks.
 {\bf (c)} Observed data matrix based on occlusion pattern. {\bf (d)} Observed data matrix with data removed randomly.}
%Best reconstruction of point tracks
%(RMSE: $0.6174$)}
\label{fig:sphere_dataset}
\end{figure}
\begin{figure}[t]
\begin{center}
\hspace{-5pt}
\begin{subfigure}{4cm}
\includegraphics[width=4cm]{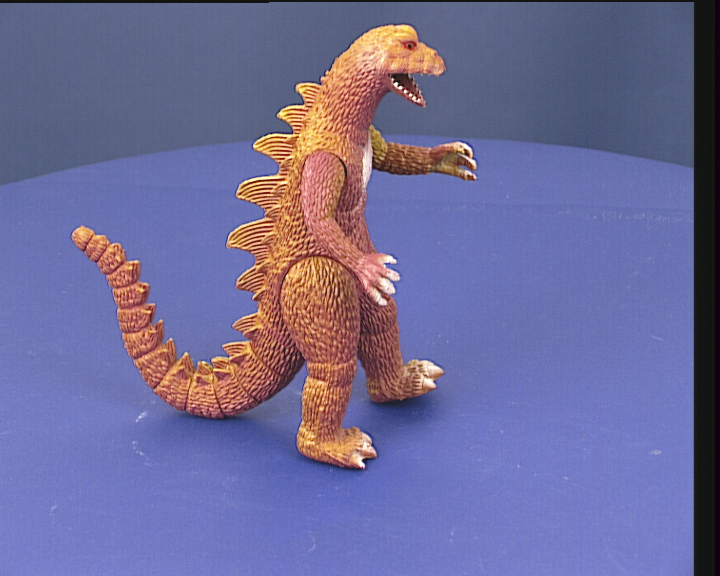}
\caption{}
\end{subfigure}
\begin{subfigure}{4cm}
\includegraphics[width=4cm]{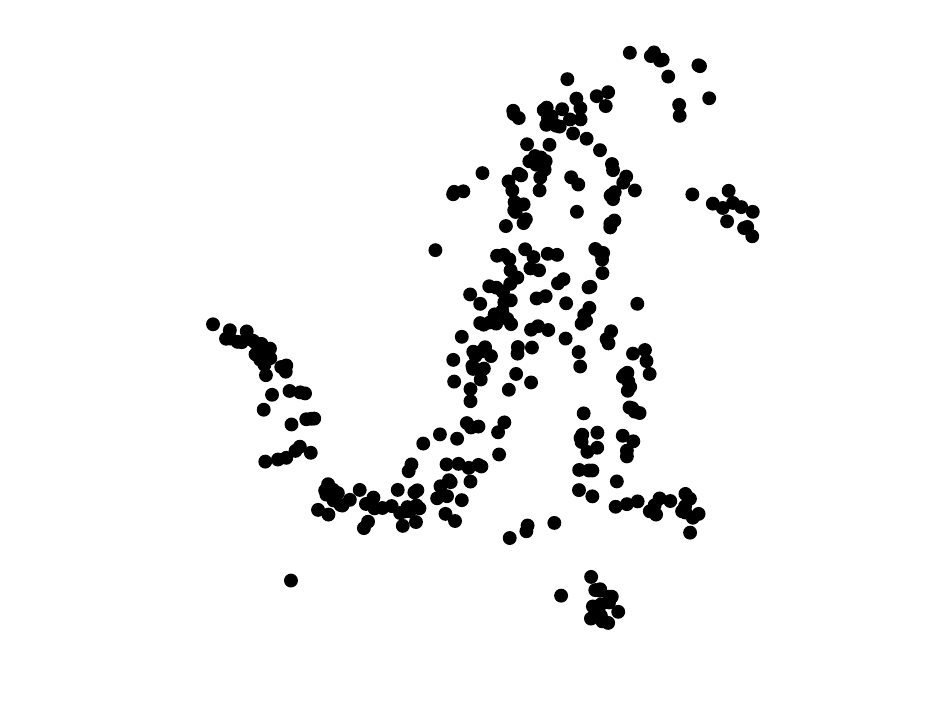}
\caption{}
\end{subfigure}
\hspace{-15pt}
\begin{subfigure}{4.6cm}
\includegraphics[width=4.6cm]{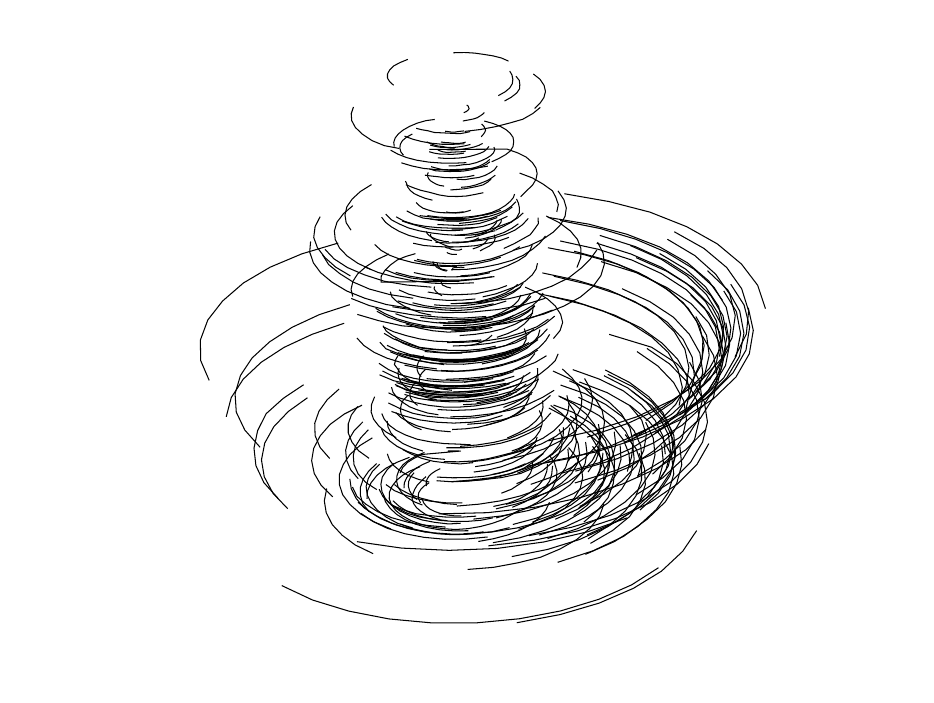}
\caption{}
\end{subfigure}
\hspace{-25pt}
\begin{subfigure}{4.6cm}
\includegraphics[width=4.6cm]{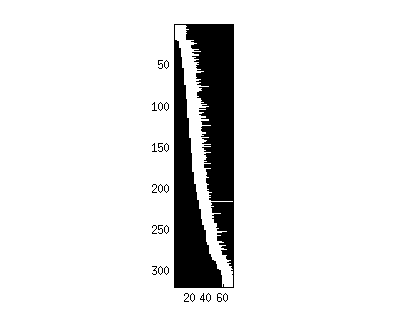}
\caption{}
\end{subfigure}
\end{center}
\caption{{\bf Dinosaur dataset}. {\bf (a)} One frame from the sequence. {\bf (b)} Reconstructed
3D model. {\bf (c)} Observed point tracks. Because the dinosaur is on a turntable, the tracks 
are elliptical. {\bf (d)} Observed data matrix.}
%Best reconstruction of point tracks (RMSE: $1.2702$).}
\label{fig:dino_dataset}
\end{figure}
\begin{figure}[t]
\begin{center}
\hspace{-5pt}
\begin{subfigure}{4cm}
\includegraphics[width=4cm]{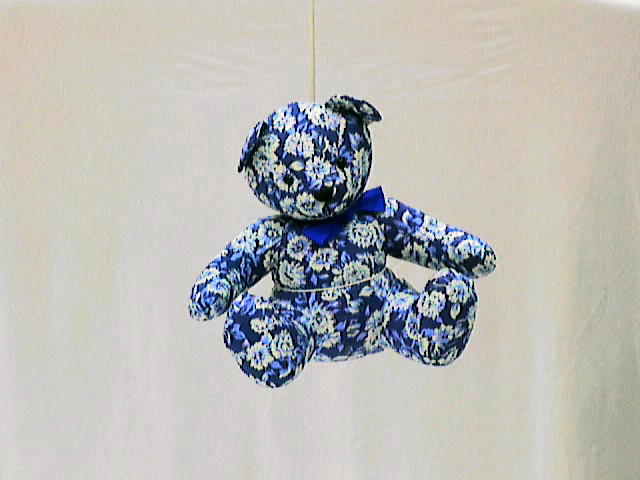}
\caption{}
\end{subfigure}
\begin{subfigure}{4cm}
\includegraphics[width=4cm]{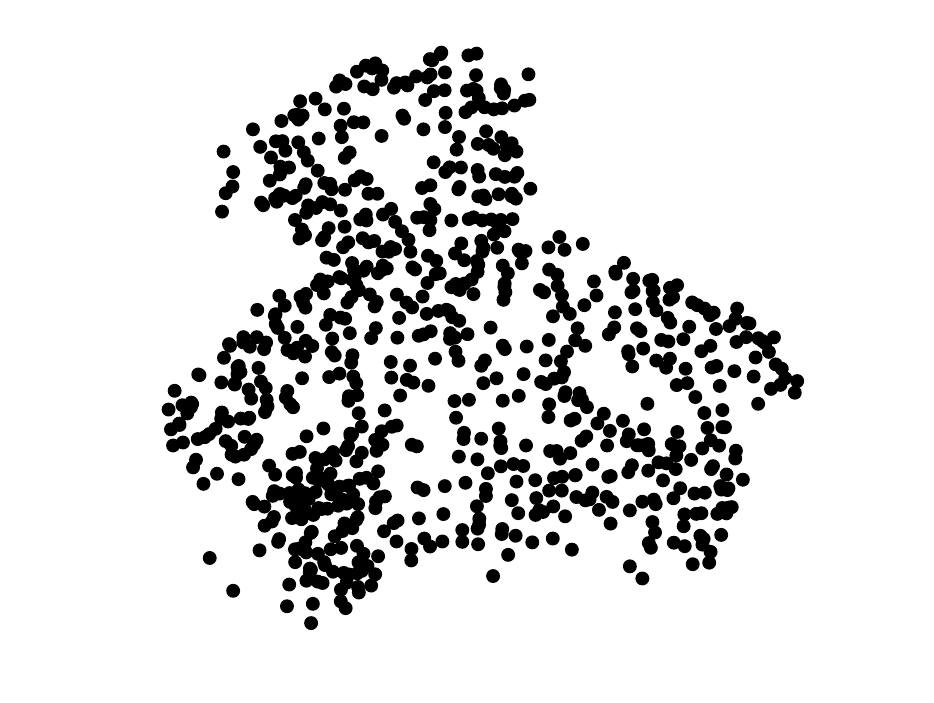}
\caption{}
\end{subfigure}
\hspace{-15pt}
\begin{subfigure}{4.6cm}
\includegraphics[width=4.6cm]{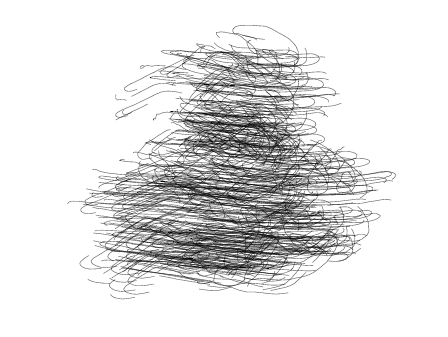}
\caption{}
\end{subfigure}
\hspace{-25pt}
\begin{subfigure}{4.6cm}
\includegraphics[width=4.6cm]{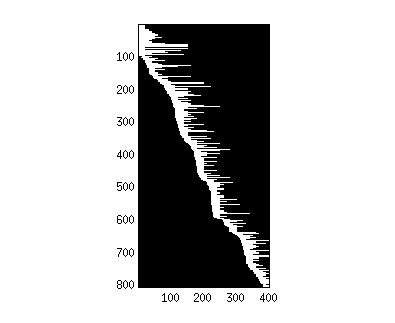}
\caption{}
\end{subfigure}
\end{center}
\caption{{\bf Bear dataset}. {\bf (a)} One frame from the sequence. {\bf (b)} Reconstructed
3D model. {\bf (c)} Observed point tracks. {\bf (d)} Observed data matix.}
%Best reconstruction of point tracks
%(RMSE: $0.6174$)}
\label{fig:bear_dataset}
\end{figure}

\subsubsection{Datasets}
Three datasets are used in our experiments: Synthetic Sphere
(Figure~\ref{fig:sphere_dataset}), Dinosaur
\cite{fitzgibbon1998automatic} (Figure~\ref{fig:dino_dataset}), and
Bear \cite{tardif2007algorithms} (Figure~\ref{fig:bear_dataset}).

The Synthetic Sphere dataset is a 3D unit sphere with $100$
randomly-placed points tracked over $200$ frames. The camera was made
to move in a smooth trajectory around the sphere and the data were
projected onto the camera using an orthographic projection, so the
problem has an exact rank-4 solution. In this case, points that
became occluded and re-appeared were treated as two separate points
and the resulting matrix is of size $143\times 400$.  We used two
different data models for this dataset. In the first, occluded data
points were removed from the data matrix, resulting in
$65.1\%$ missing data with a banded structure to the data matrix.  In
the second, we removed the same number of data points uniformly at
random.

The Dinosaur (``Dino'') dataset consists of $319$ points over $36$
frames, with $76.9\%$ missing data. The Bear dataset has $806$ points
and $200$ frames, with $88.6\%$ missing data.

The Synthetic Dino dataset was generated from the Dino sequence as
follows. The 3D model of the best reconstruction of the Dino was used
as the ground truth, and was projected onto $100$ random orthographic
cameras, resulting in a matrix of size $319\times 200$. $90\%$ of the
data were then removed uniformly at random.

\subsubsection{Results for the Synthetic Sphere}

\begin{figure}[h!]
\centering
\begin{subfigure}{16cm}
\includegraphics[width=\textwidth]{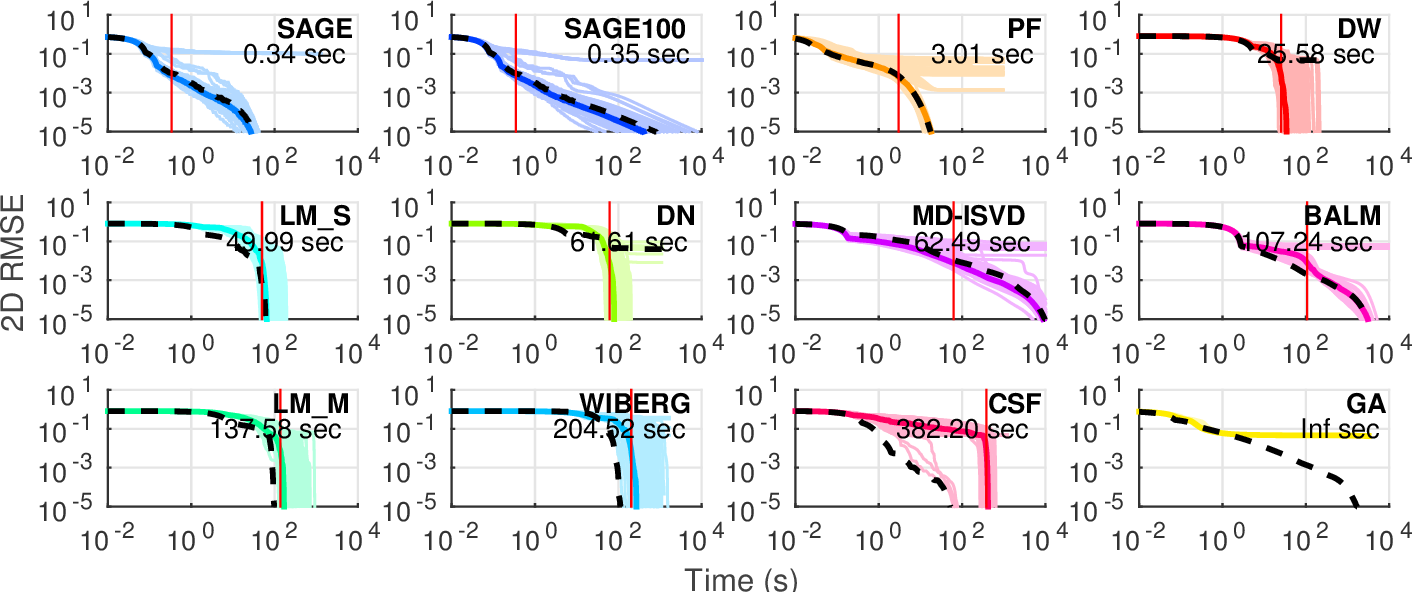}
\label{fig:batch_dino_all}
\vspace{-10pt}
\end{subfigure}
\begin{subfigure}{16cm}
\includegraphics[width=\textwidth]{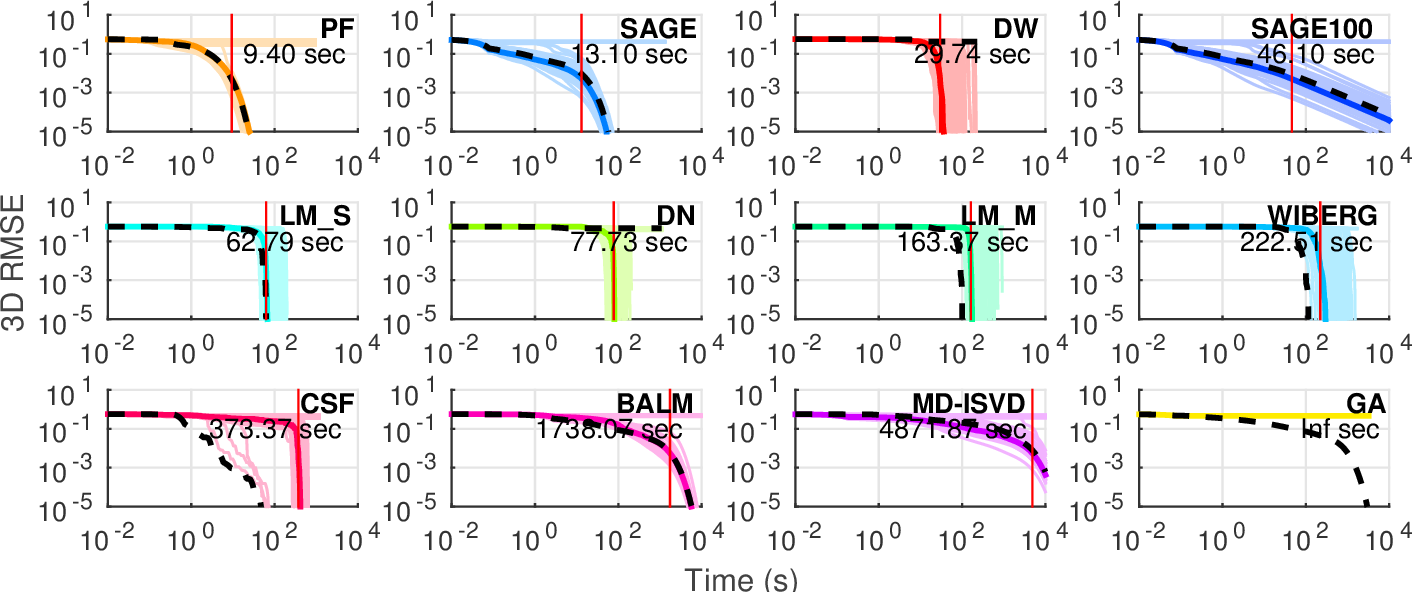}
\end{subfigure}
\caption{Results on the Synthetic Sphere dataset with a banded
  occlusion pattern. {\bf Top plots} show 2D RMSE over time. All 100
  random initializations are plotted in light colors, the median of
  which is plotted in a darker color. The result of the deterministic
  initialization is shown as a dashed black line. The number below
  each algorithm name is the time for the median run to go 99\% of the
  way from the initial error to the minimum error.  The location of
  this point is shown using a vertical red line and the algorithms are
  sorted by this number. {\bf Bottom plots} show the same values, but
  for 3D RMSE. }
\label{fig:sphere1}
\end{figure}

\begin{figure}[h!]
\centering
\begin{subfigure}{16cm}
\includegraphics[width=\textwidth]{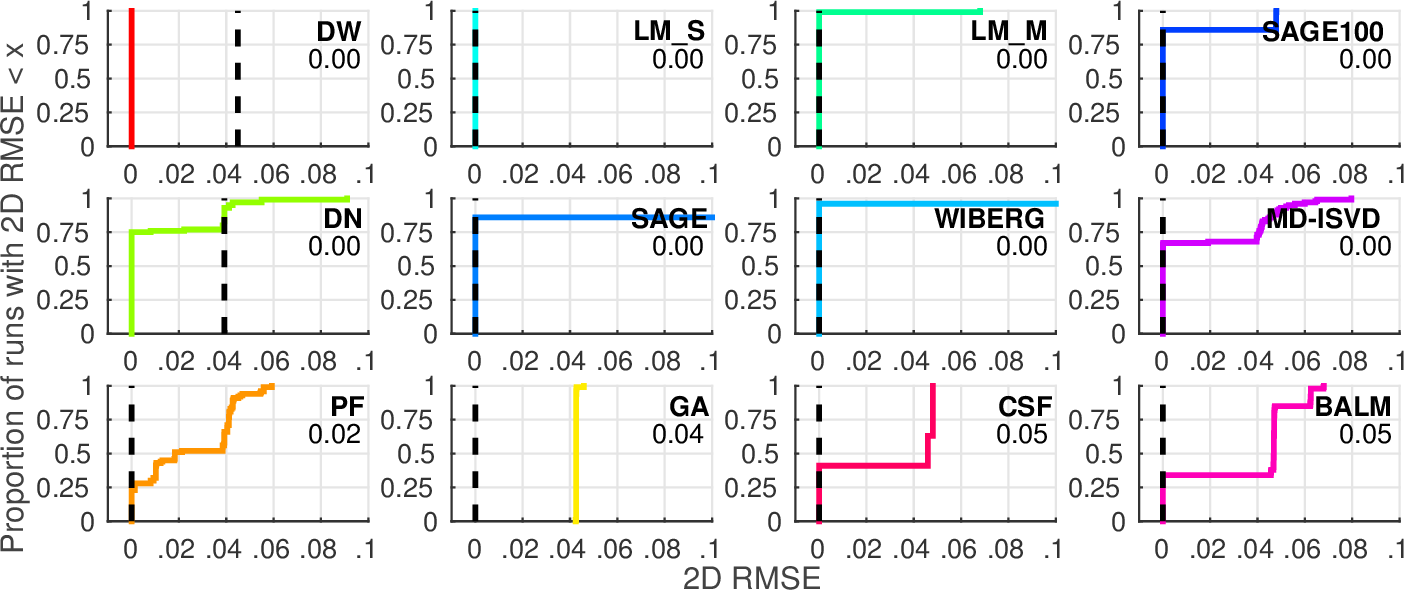}
\end{subfigure}
\caption{Empirical CDF results on the Synthetic Sphere dataset with a banded
  occlusion pattern. These plots show the empirical cumulative
  distribution of error after convergence. The error of the
  deterministic initialization is shown as a vertical dashed black
  line. The number associated with each algorithm is the median final
  error, and the algorithms are sorted by this value. The CDF plots
  for 3D RMSE are very similar and thus are not shown.}
\label{fig:sphere1cdf}
\end{figure}

{\bf Banded Occlusion Pattern}\quad We begin with the
  Synthetic Sphere dataset with a realistic occlusion pattern. The
  results of measuring the 2D RMSE for this dataset are shown in
  Figure \ref{fig:sphere1}. The top set of plots show the convergence
  of each algorithm over time.  We plot all 100 randomly-initialized
  runs in a lighter color, and the median run is plotted in a darker
  version of the same color.  The dashed black line shows the run
  using the deterministic initialization. To measure the speed of
  convergence, we find the time it takes for the median line to move
  99\% of the way from its initial error to the smallest error
  achieved over all plots. The algorithms are sorted by this value and
  its location is denoted by a vertical red line.  The bottom set of
  plots shows the same results when 3D RMSE is calculated.  The 
  set of plots in Figure \ref{fig:sphere1cdf} show the empirical cumulative distribution function of
  the error for each algorithm after convergence. Once again, the
  error for the deterministic initialization is shown with a black
  dashed line. In this case, the number associated with each algorithm is the median error
  after convergence and the algorithms are sorted by this value.

In terms of speed, we find that SAGE and SAGE100 are
  significantly faster than any other algorithm for 2D RMSE. These two
  algorithms take only 0.35 seconds to achieve 99\% of the final
  error reduction. The only algorithm with comparable performance is
  PF, which takes 10 times longer and does not converge as
  frequently. All other algorithms are at least two orders of magnitude slower than
  SAGE. It is also notable that the deterministic solution often
  performs as good as the best random initialization, with the
  exceptions being DW and DN. A similar result is seen with 3D RMSE in the
  middle plots, where PF and SAGE converge much faster than
  other algorithms. The general convergence rate in terms of 3D error
  is lower than 2D error, since the 3D error will generally not
  decrease much until the 2D error is lowered enough for an accurate 3D
  model to be generated.
  
It is important to note here that while SAGE reaches a small
  error in computation times that are orders of magnitude faster than
  the other algorithms, it is not always fastest to reach a
  \emph{very} small error. Consider the time it takes to reach
  $10^{-2}$ versus $10^{-5}$ for 2D RMSE in this experiment. While
  SAGE reaches $10^{-2}$ in a median time of 0.26 seconds, it takes an additional
  26.9 seconds to reach $10^{-5}$.  Damped Wiberg (DW), by
  comparison, reaches  $10^{-2}$ in 25.1 seconds and takes only another 9.52 seconds
  to reach $10^{-5}$. Thus, while our first-order method is
  faster at first, second-order methods have potential to reach a
  highly accurate solution, given more computation time.  Since our
  focus is real-time structure from motion, we believe that SAGE is an
  excellent option for achieving aceptable accuracy in significantly
  less time.

With respect to convergence as shown in the bottom set of
  plots, we find that LM\_M and LM\_S almost always converge,
  regardless of their initialization; the same can be said for DW if
  we ignore the deterministic initialization. Other algorithms, including SAGE
  and WIBERG, also converge in most cases.  Interestingly, the more
  complex algorithms CSF and BALM will often get stuck in local minima
  on this dataset. We note that these algorithms have several extra
  parameters that affect their convergence rate and it may be possible
  to tune them to specific datasets to get better convergence.

\bigskip

\begin{figure*}[h!]
\centering
\begin{subfigure}{16cm}
\includegraphics[width=\textwidth]{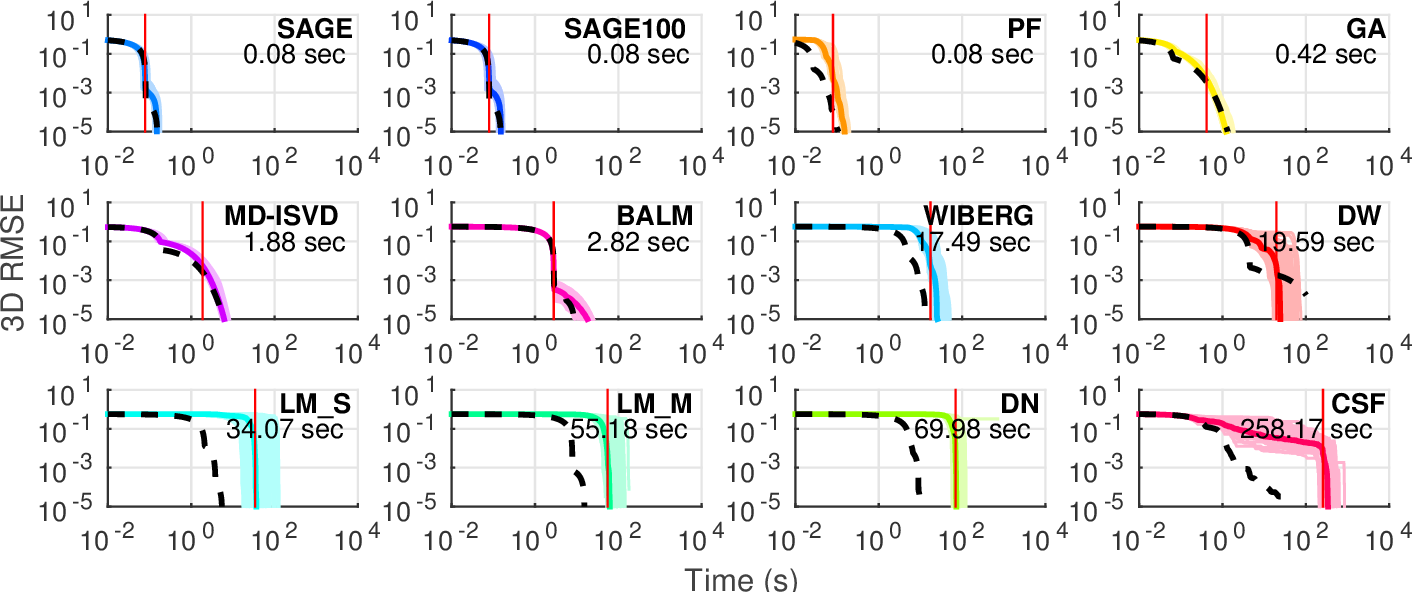}
\end{subfigure}
\caption{Results on the synthetic Sphere dataset with a random occlusion pattern, 
measured using 3D RMSE. See the text for details.
All algorithms achieve nearly universal convergence on this dataset, and we find that 
SAGE, SAGE100 and PF are much faster than other approaches.}
\label{fig:sphere2}
\end{figure*}

{\bf Random occlusion pattern.}\quad Results for the Synthetic Sphere
dataset using a random pattern of missing data are shown in
Figure~\ref{fig:sphere2}.  This dataset is much simpler for all
algorithms; 100\% convergence is achieved for all initializations.  We
omit 2D errors since they are nearly identical to the 3D error shown
here. We omit the CDF plots too, since convergence is universally
achieved. Again, we find that SAGE, SAGE100 and PF are several orders
of magnitude faster than most other algorithms. For this dataset,
these three algorithms converge to the groundtruth 3D sphere in less
than a tenth of a second. Interestingly, in this case there
  is no accuracy trade-off; SAGE reaches an error as low as the best
  algorithm in orders of magnitude less computation time.

%\begin{figure}[H]
%\centering
%\begin{subfigure}{16cm}
%\includegraphics[width=\textwidth]{sphere1_all_2D.png}
%\label{fig:batch_dino_all}
%\vspace{-10pt}
%\end{subfigure}
%\begin{subfigure}{16cm}
%\includegraphics[width=\textwidth]{sphere1_all_3D.png}
%\end{subfigure}
%\begin{subfigure}{16cm}
%\includegraphics[width=\textwidth]{sphere1_cdf_2D.png}
%\end{subfigure}
%\caption{Results on the Synthetic Sphere dataset with a banded
%  occlusion pattern. {\bf Top plots} show 2D RMSE over time. All 100
%  random initializations are plotted in light colors, the median of
%  which is plotted in a darker color. The result of the deterministic
%  initialization is shown as a dashed black line. The number below
%  each algorithm name is the time for the median run to go 99\% of the
%  way from the initial error to the minimum error.  The location of
%  this point is shown using a vertical red line and the algorithms are
%  sorted by this number. {\bf Middle plots} show the same values, but
%  for 3D RMSE.  {\bf Bottom plots} show the empirical cumulative
%  distribution of error after convergence. The error of the
%  deterministic initialization is shown as a vertical dashed black
%  line. The number associated with each algorithm is the median final
%  error, and the algorithms are sorted by this value. The CDF plots
%  for 3D RMSE are very similar and thus are not shown.}
%\label{fig:sphere1}
%\end{figure}

%The other algorithms -- PF, GA and DN -- 
%rarely if ever converge to the best solution. 

\subsubsection{Results for the Dino and Bear}

\begin{figure*}[h!]
\centering
\begin{subfigure}{16cm}
\includegraphics[width=\textwidth]{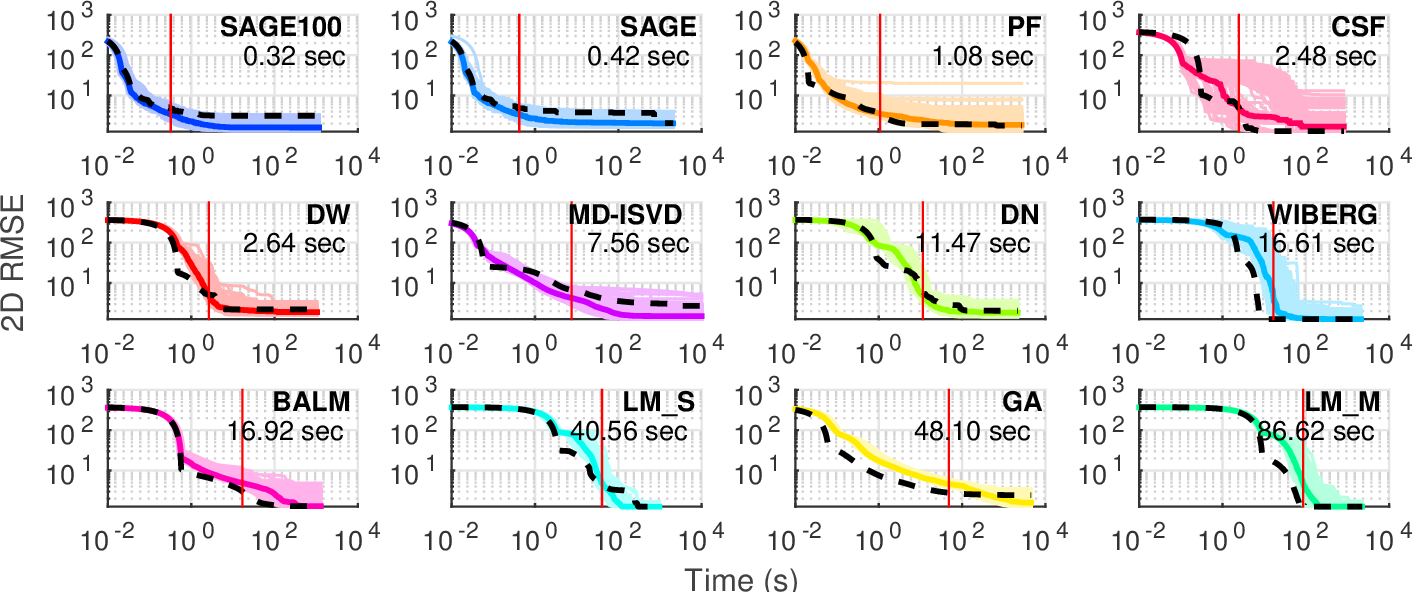}
\label{fig:batch_dino_all}
\end{subfigure}
\begin{subfigure}{16cm}
\includegraphics[width=\textwidth]{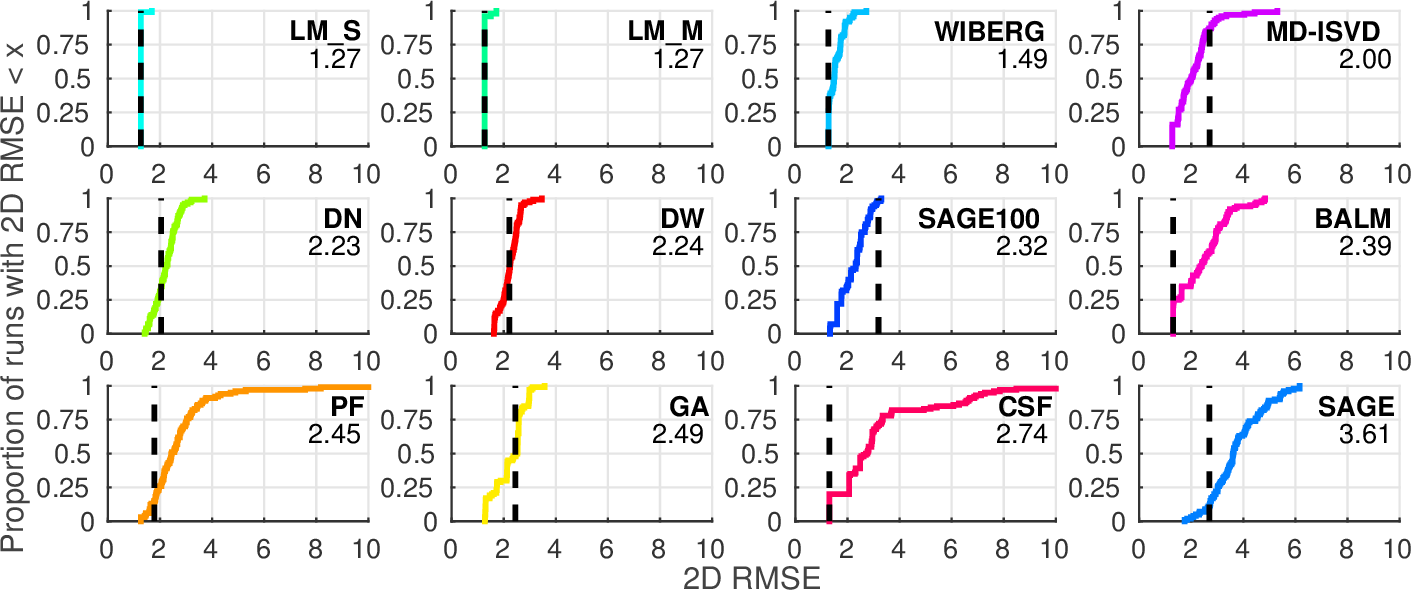}
\end{subfigure}
\caption{Result for the Dino dataset. {\bf Top plots} show the convergence of algorithms over time. 
{\bf Bottom plots} show the empirical cumulative distribution of 2D RMSE for each algorithm. In both cases,
the deterministic initialization is shown using a dashed black line. See the text for details.}
\label{fig:batch_dino}
\end{figure*}

\begin{figure*}[h!]
\centering
\begin{subfigure}{16cm}
\includegraphics[width=\textwidth]{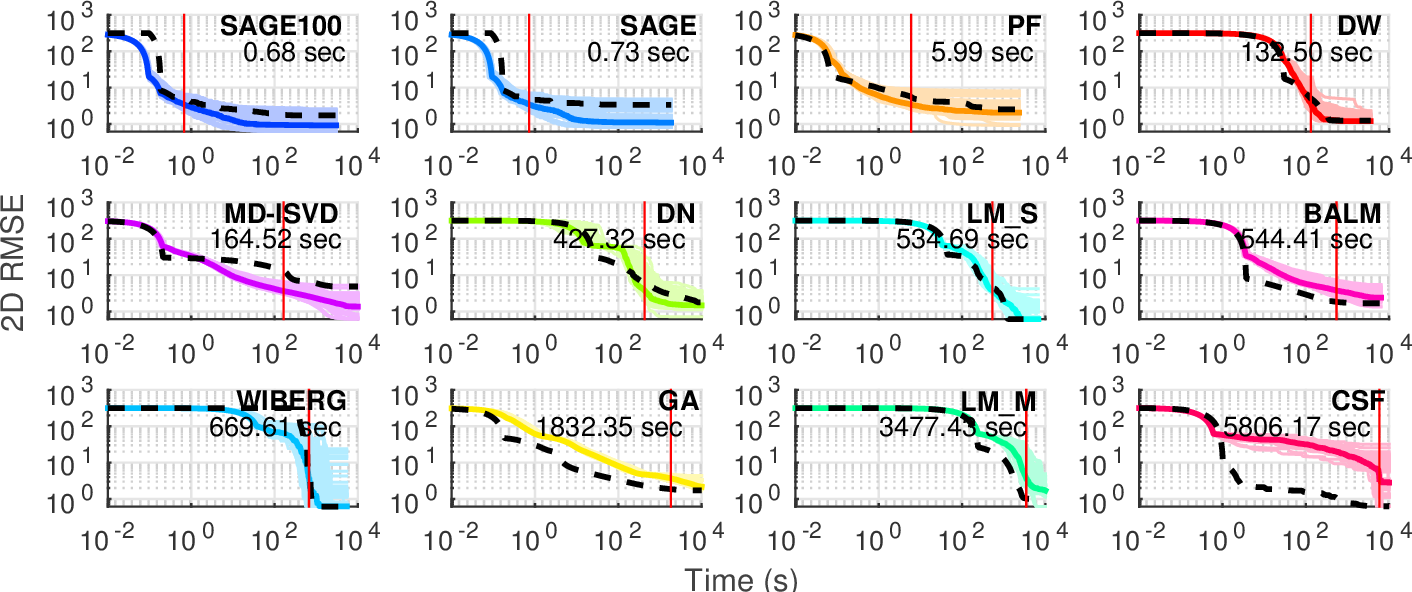}
\label{fig:batch_bear_all}
\end{subfigure}
\begin{subfigure}{16cm}
\includegraphics[width=\textwidth]{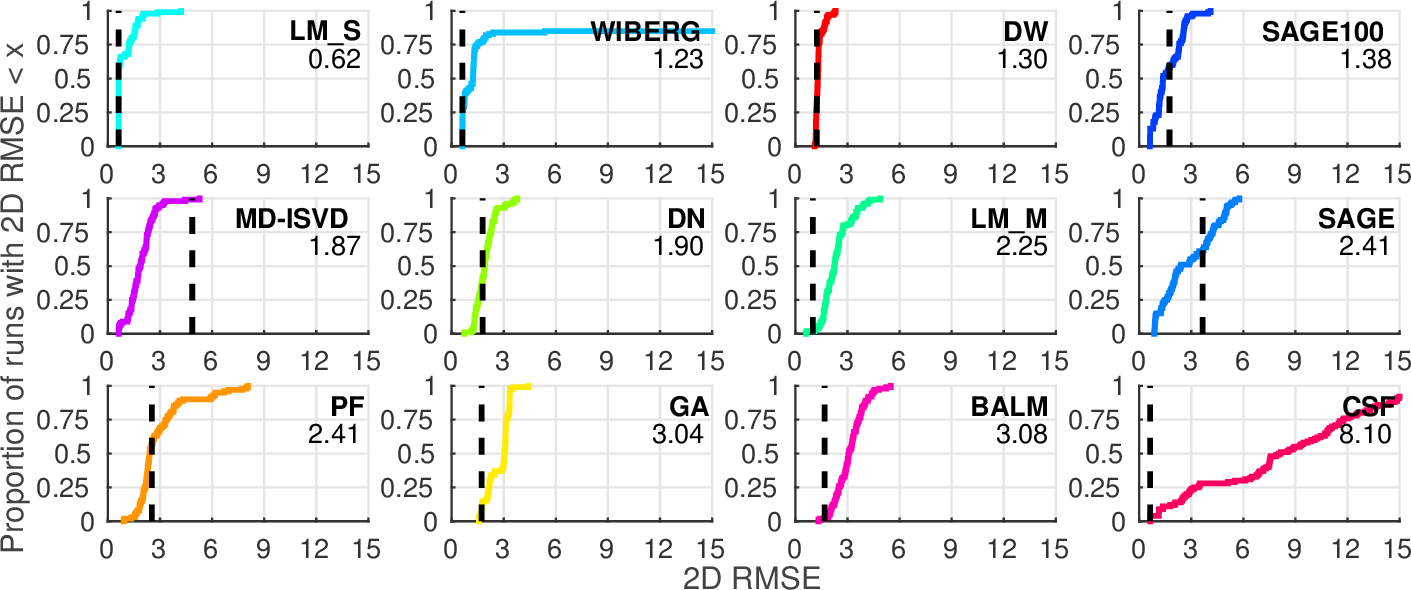}
\end{subfigure}
\caption{Result for the Bear dataset. {\bf Top plots} show the convergence of algorithms over time. 
{\bf Bottom plots} show the empirical cumulative distribution of 2D RMSE for each algorithm. In both cases,
the deterministic initialization is shown using a dashed black line. See the text for details.}
\label{fig:batch_bear}
\end{figure*}

Results for the Dino and Bear datasets are shown in
Figures~\ref{fig:batch_dino} and \ref{fig:batch_bear},
respectively. For both datasets, the results are relatively
similar. We find that SAGE100 and SAGE are again much faster than all
other algorithms, with the scaled version SAGE100 being moderately
faster than the unscaled SAGE. In terms of convergence, we again see
that the second-order methods LM\_S and LM\_M converge nearly every
time, with WIBERG also performing well.  We also find the SAGE100
converges more often than the unscaled SAGE.  For the Bear dataset,
both CSF and BALM exhibit slow convergence, because the dataset is
larger. We may have observed more frequent convergence if we had
allowed more computation time, but our upper limit of $10^4$ seconds is a generous time
limit, and if the algorithm can not solve the problem in this time, it
is worth looking for alternatives.

The trade-off between accuracy and computation time is even
  more prominent in these real datasets. Several algorithms converge
  to a solution that is not as accurate as the most accurate of the
  slower methods. One conclusion we draw is that if an accurate batch
  solution is required, it may be beneficial to combine SAGE100 with a
  second order method, in a two-phase strategy. SAGE100 would achieve
  a moderately accurate solution quickly in the first phase, while the
  second order method finds a highly accurate final solution in the
  second phase.

\subsubsection{Robust algorithms}
\begin{figure*}[t]
\begin{center}
\begin{subfigure}{8.5cm}
\includegraphics[width=8.5cm]{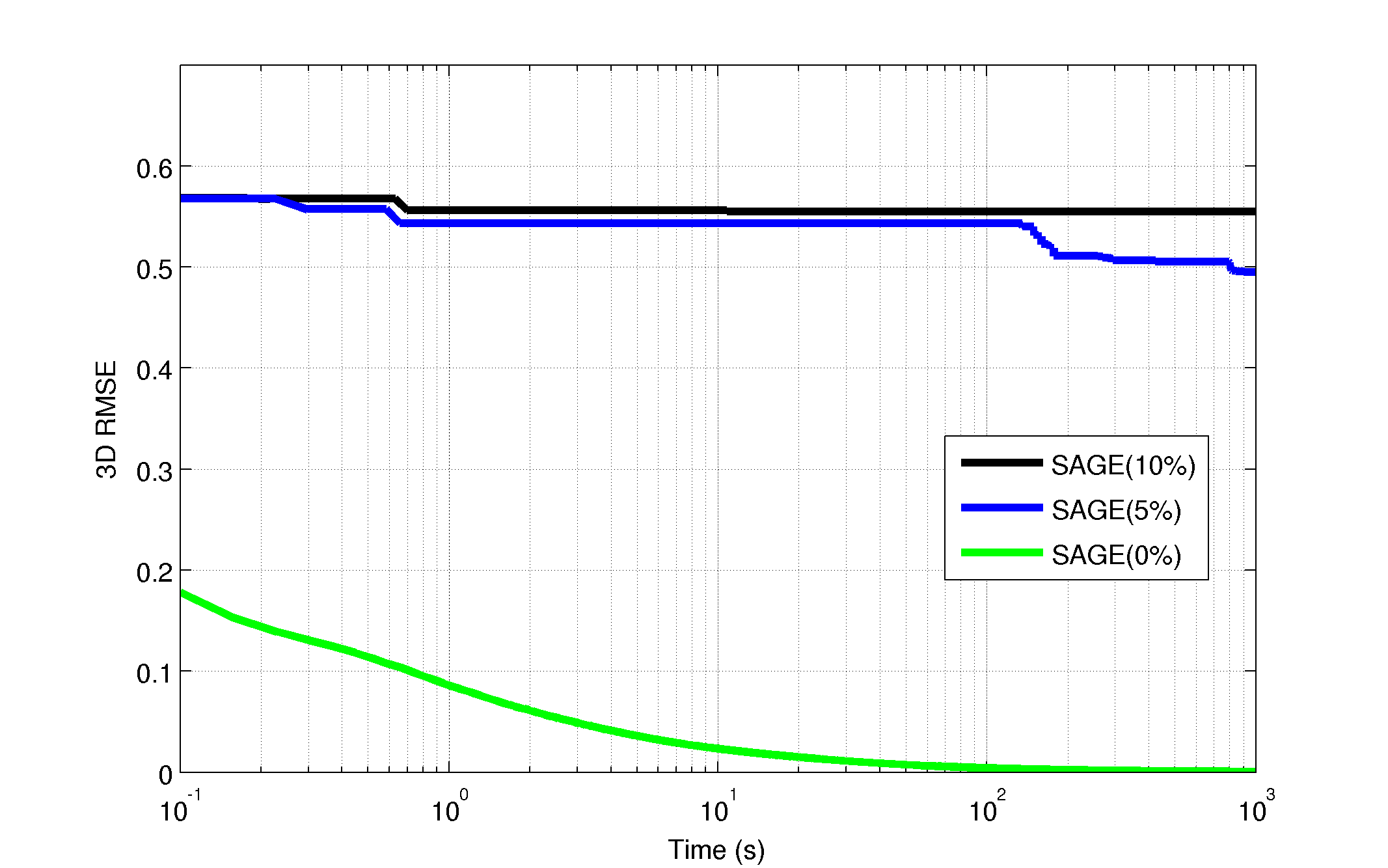}
\caption{SAGE}
\end{subfigure}
\hspace{-20pt}
\begin{subfigure}{8.5cm}
\includegraphics[width=8.5cm]{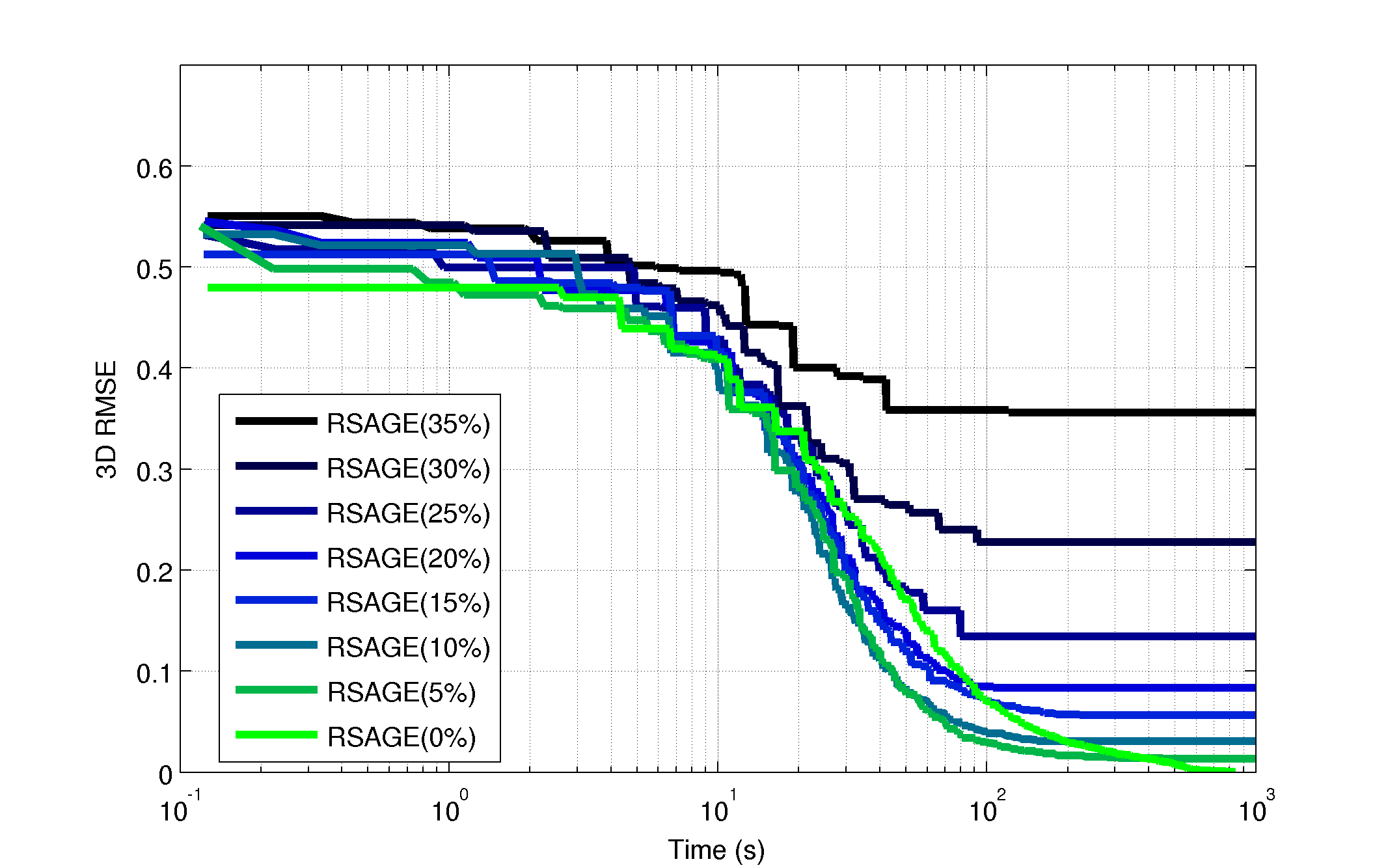}
\caption{RSAGE100}
\end{subfigure}
\end{center}
\caption{Effect of sparse noise on SAGE100 and RSAGE100 using the Synthetic Sphere dataset with
a banded occlusion pattern.
The algorithms have a decay factor on the residual of $\alpha_t=C/(C+t)$,  with $C=100$.
A variable proportion of the data were set to sparse outliers, from $0\%$ to $35\%$,
and we measure the 3D RMSE over time.
Even a small number of sparse outliers causes SAGE to converge to
an  incorrect 3D model, while the 3D RMSE decays gradually with RSAGE.
When no sparse outliers are present, SAGE converges much faster than RSAGE.}
\label{fig:batch_robust}
\end{figure*}

We compare SAGE100 with its robust counterpart RSAGE100, with results
shown in Figure~\ref{fig:batch_robust}. We used the Synthetic Sphere
dataset with a banded occlusion pattern.  A varying proportion of the
visible entries --- from $0\%$ to $35\%$ --- were set to sparse
outliers in the range $[-100,100]$.  Both SAGE100 and RSAGE100 were
initialized using a deterministic initialization by filling in missing
values with the mean of all values present in each column.

With even a small number of sparse outliers, SAGE is unable to find
the correct 3D model.  RSAGE handles the sparse outliers much better,
with the 3D RMSE decaying gradually as the proportion of sparse
outliers increases.  For SFM problems containing sparse outliers,
RSAGE can therefore be used to find an accurate 3D model.

We note that there are many batch approaches designed
  specifically for robust matrix factorization in the presence of
  sparse outliers
  \cite{eriksson2010efficient,zheng2012practical,strelow2012general}. The
  purpose here is only to note that robustness can be added to our own
  model in a straightforward way.
%  leading to an efficient, robust online approach to structure from
%  motion.
We leave a thorough comparison to robust batch methods as future
work.  Since we have found RSAGE to be
much slower than SAGE,
%  as can be seen in the convergence rates of the two methods when no
% sparse outliers are present.  Thus, for applications which require
real-time performance demands may require us to to adjust point
tracking so that only high-quality tracks are used, and run SAGE
(rather than RSAGE) on this reduced data set.

\subsubsection{Large dataset}
In addition to its speed, another advantage of SAGE is that it is memory efficient and can be used
on very large datasets where second-order methods are much too computationally burdensome. In \cite{del2012bilinear},
the algorithm BALM was used to reconstruct a very large synthetic 3D model, and we
compare SAGE to them using the same dataset. The dataset used is Venice from Agarwal et.~al \cite{agarwal2010bundle}, as modified in \cite{del2012bilinear}:  the 3D model was projected onto $100$ random orthographic cameras, producing a measurement matrix of size $939551\times 200$ from which $90\%$ of the entries were randomly removed.  We measure the error as $\|S-S_{gt}\|_F/\|S_{gt}\|_F$, where $S$ is the resulting 3D reconstruction and $S_{gt}$ is the ground-truth 3D model. The algorithm of \cite{del2012bilinear} reported a reconstruction error of $6.67\%$, with no computational time reported. SAGE achieved this reconstruction error within $50$ seconds. SAGE then further reduced the error to $2.48\%$ after $2$ minutes, and to  $1.37\%$ after $10$ minutes.

\subsection{Online experiments} \label{sec:online}
\subsubsection{Real-time implementation}
\label{sec:real_time}
To demonstrate the use of our method in real-time reconstruction, 
we implemented SAGE in C++ using OpenCV.
Our implementation uses two threads running on separate cores: One thread reads frames from
an attached webcam and tracks point using the KLT tracker in OpenCV, while the other thread continually
runs SAGE and incorporates new data as it becomes available. We used a MacBook Pro
with a 2.66 GHz Intel Core 2 Duo processor and 8 GB of memory, with a Logitech C270 webcam.

The implementation was run at $15$ fps and we used it to capture and reconstruct a 3D model of 
a toy giraffe in real time (Figure~\ref{fig:giraffe_dataset}). The bottleneck in this process is tracking points between frames, and during this time SAGE processed an average of $205$ columns per frame, each column
being selected randomly from
the past columns of the measurement matrix.
Since this approach is able to use a simple webcam and a standard computer
to obtain a 3D model in real time, we envision it 
 being useful in applications where it is desirable to build a 3D model with low-cost hardware, such as
at-home 3D printing.

\subsubsection{Comparison}
In the online setting, we compare MD-ISVD to SAGE for which the residual is not scaled, as well as to SAGE100
for which the residual for a column $c$ is scaled by $C/(C+t(c))$ where $t(c)$ indicates the number
of times that column $c$ has been processed. 
A related incremental algorithm is also given in \cite{cabral2011fast}. We attempted to compare to this method as well,
but found that it was extremely unstable because it requires calculating the determinant of matrices with
very large eigenvalues. For this reason, its results are not presented here.
Each algorithm
was implemented in MATLAB for this comparison.

In contrast to the batch case, for the online experiments
the frames are processed sequentially and only data up to the current frame is available. The algorithms
are compared by varying the number of iterations spent processing each frame and evaluating the RMSE
after the last frame.

\begin{figure*}[t]
\begin{center}
\hspace{-5pt}
\begin{subfigure}{4cm}
\includegraphics[width=4cm]{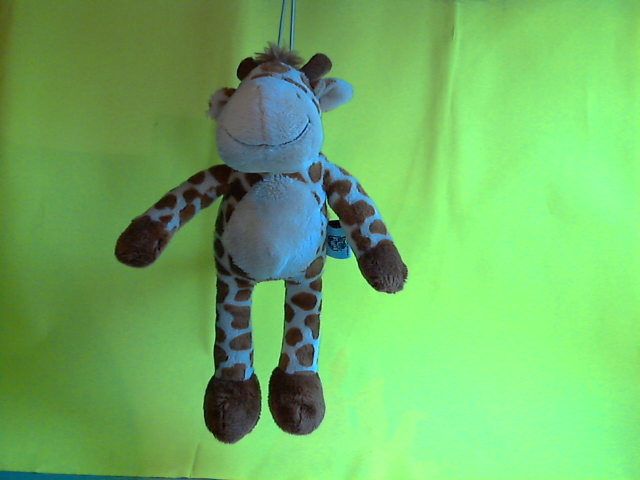}
\caption{}
\end{subfigure}
\begin{subfigure}{4cm}
\includegraphics[width=4cm]{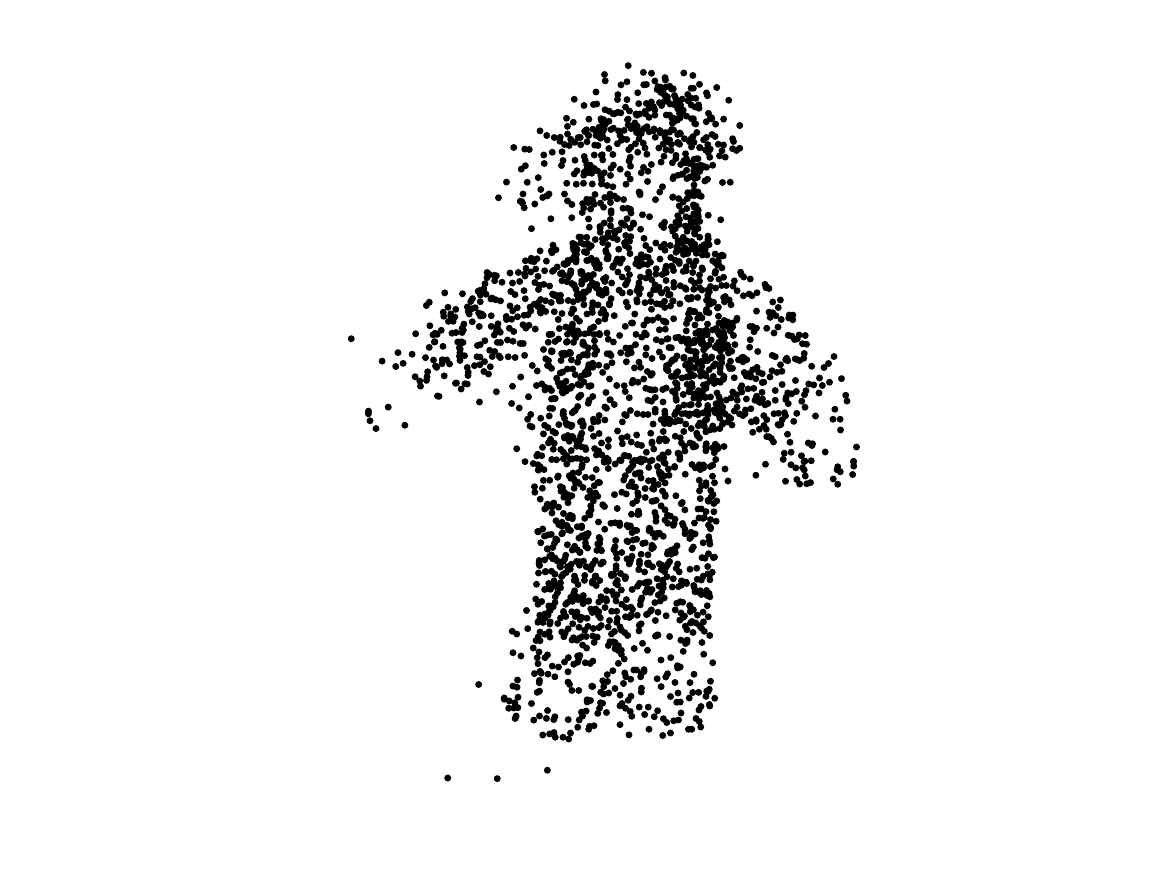}
\caption{}
\end{subfigure}
\hspace{-15pt}
\begin{subfigure}{4.6cm}
\includegraphics[width=4.6cm]{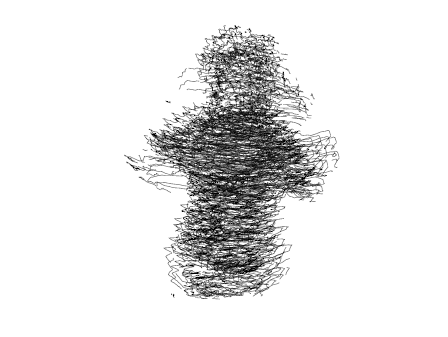}
\caption{}
\end{subfigure}
\hspace{-25pt}
\begin{subfigure}{4.6cm}
\includegraphics[width=4.6cm]{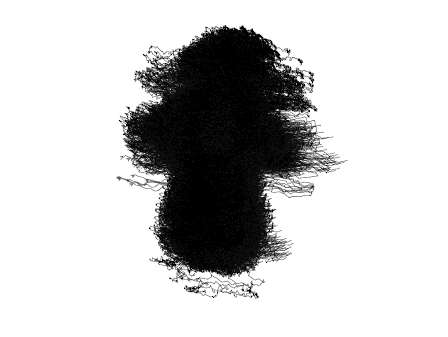}
\caption{}
\end{subfigure}
\end{center}
\caption{{\bf Giraffe dataset}. {\bf (a)} One frame from the sequence. {\bf (b)} Reconstructed
3D model. {\bf (c)} Observed point tracks. {\bf (d)} Final reconstruction of point tracks after running SAGE online.}
\label{fig:giraffe_dataset}
\end{figure*}

In addition to the Sphere, Dino, and Bear sequences, we also use a new sequence of a stuffed Giraffe. 
This dataset was gathered in real time; the details of the real time implementation are given in Section \ref{sec:real_time}.
The Giraffe dataset is shown in Figure~\ref{fig:giraffe_dataset}, and consists of $2634$ points over $343$ frames, with $93.4\%$ missing data. Note that
this dataset is significantly larger than the other sequences and most second-order
batch algorithms would be computationally intractable due to the large number of variables.

\begin{figure}[h]
\begin{center}
\begin{subfigure}{8.0cm}
\includegraphics[width=8.0cm]{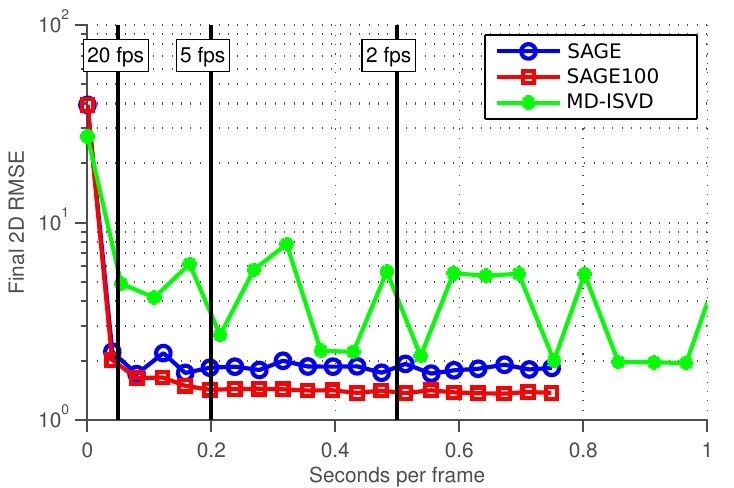}
\caption{\bf Dino}
\end{subfigure}
\begin{subfigure}{8.0cm}
\includegraphics[width=8.0cm]{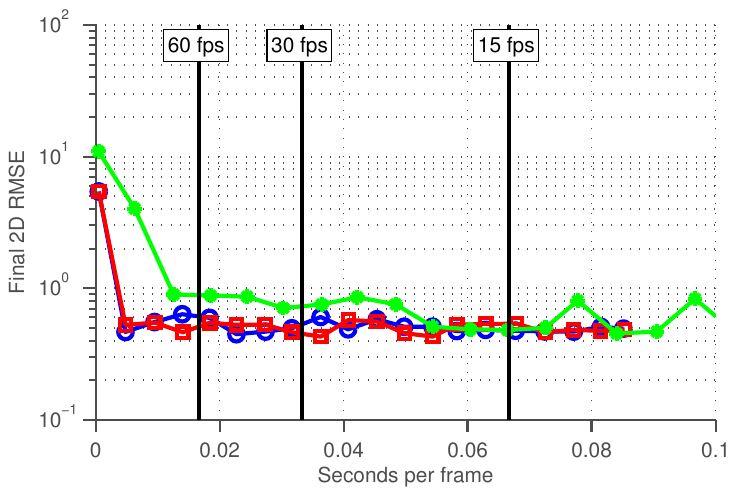}
\caption{\bf Bear}
\end{subfigure}
\\
\begin{subfigure}{8.0cm}
\includegraphics[width=8.0cm]{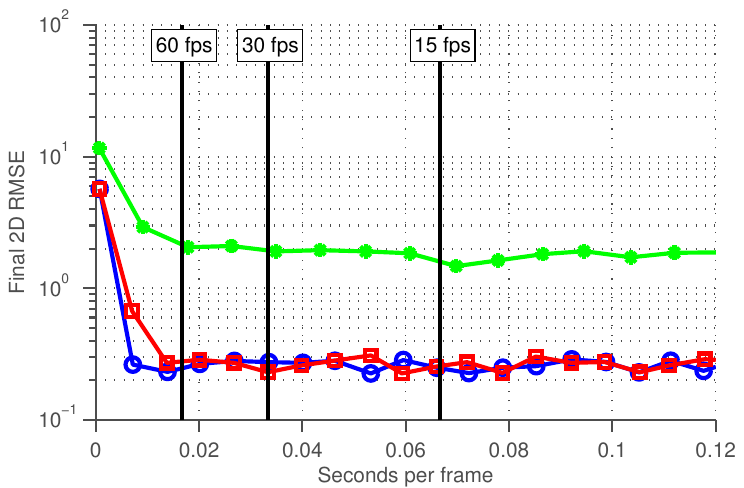}
\caption{\bf Giraffe}
\end{subfigure}
\begin{subfigure}{8.0cm}
\includegraphics[width=8.0cm]{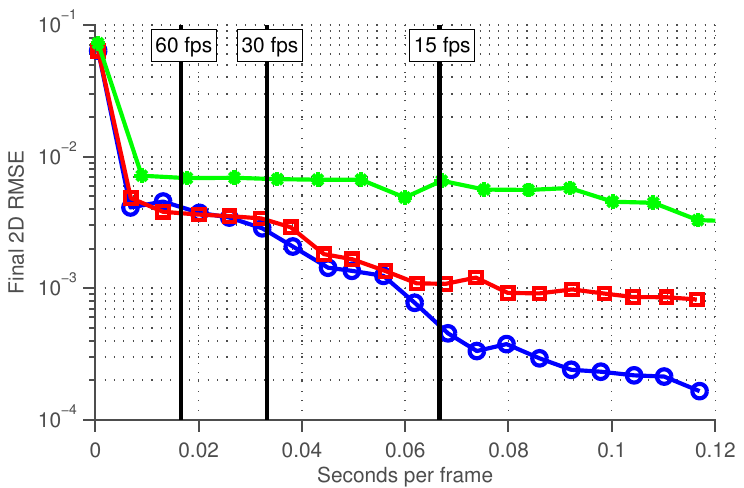}
\caption{\bf Sphere (2D RMSE)}
\end{subfigure}
\begin{subfigure}{8.0cm}
\includegraphics[width=8.0cm]{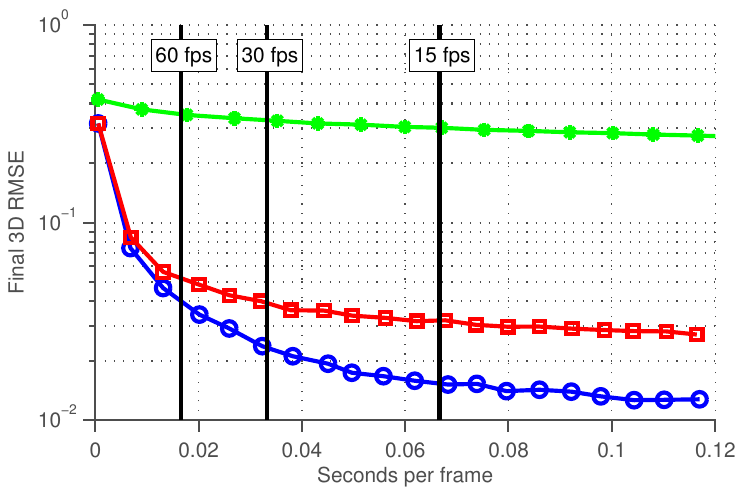}
\caption{\bf Sphere (3D RMSE)}
\end{subfigure}
\begin{subfigure}{7.0cm}
\includegraphics[width=7.0cm]{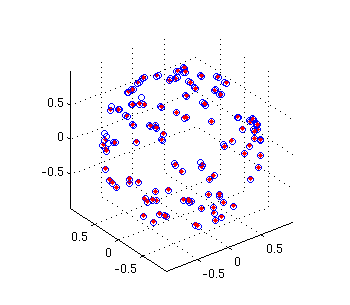}
\caption{\bf Groundtruth 3D model (red) and estimated model (blue) for SAGE after running at 15 fps.}
\end{subfigure}
\end{center}
\caption{{\bf Online experiments}. We compare MD-ISVD, SAGE and SAGE100 and vary the number
of iterations spent on each frame. In all cases, SAGE and SAGE100 achieve a low RMSE
even for a high framerate. For the synthetic Sphere dataset, we shown both 2D and 3D RMSE. In (f), we shown the
3D model generated using SAGE in an online setting when run at 15 frames per second on the Sphere dataset.}
\label{fig:online}
\end{figure}

Results for online experiments are given in Figure~\ref{fig:online}.  In all cases, SAGE and SAGE100 outperform
MD-ISVD over a range of framerates. In addition, we find that for the Dino sequence, the
scaled algorithm SAGE100 performs somewhat better than the unscaled SAGE; the decaying step size seems to help.
The opposite is observed for the Sphere dataset, where SAGE outperforms SAGE100, but in this case
there is no noise and so reducing downscaling the residual only serves to reduce the convergence rate.

SAGE is able to achieve
a very low RMSE in all cases using a very high framerate. For the shorter Dino sequence,
it may be necessary to reduce the framerate to $5$ fps to achieve a low RMSE, but for
the other sequences a  high framerate can be used since the algorithm has more
of an opportunity to re-visit old frames in these longer sequences. In fact, for the Bear and Giraffe sequences,
 a framerate
of $60$ fps is sufficient for an accurate reconstruction.

For the Sphere dataset, Figure~\ref{fig:online} also shows the resulting 3D reconstruction that is obtained
after running SAGE at 15 fps. The model obtained is very close to the groundtruth, demonstrating that we are
indeed able to find accurate solutions in real time.

\section{Conclusion}
In this paper we have proposed the use of SAGE and its robust
counterpart RSAGE for factorization-based structure from
motion. Although there are many other approaches for this problem,
they are generally either fast but with unreliable convergence (PF and
GA) or use second-order information to improve their convergence but
are slower (DN, LM\_M and LM\_S). We have demonstrated that SAGE
converges quickly to solutions of good quality. SAGE performs
exceptionally well on batch problems and is orders of magnitude faster
than other algorithms that achieve a similar accuracy. However, we
have also observed that SAGE can have difficulty reaching a highly
accurate best solution, due to it being a first-order stochastic
gradient algorithm.  For applications where both speed an accuracy are
required, SAGE could be used to rapidly find a solution of
reasonable quality, which can then be used as a starting point for
another algorithm that may find a more accurate final solution
(such as LM\_M or LM\_S).

We have also shown that SAGE performs well in the online setting,
where the aim is to build a 3D model of an object in real time as the
video is taken. The efficiency of SAGE allows the use of low-cost
hardware such as a laptop and webcam, or a
cellphone. Furthermore, even if a slow-but-accurate algorithm such as
LM\_M or LM\_S is used to obtain the final 3D model, the structure
determination process is sped up by having a good initialization
available from the real-time execution of SAGE.

\bigskip
\noindent
\section{Acknowledgements} 
Work on this paper by Laura Balzano was partially supported by Army
Research Office grant W911NF-14-1-0634. Stephen Wright acknowledges
the support of NSF Awards DMS-1216318 and IIS-1447449, ONR Award
N00014-13-1-0129, AFOSR Award FA9550-13-1-0138, and Subcontract
3F-30222 from Argonne National Laboratory.

%\ifCLASSOPTIONcaptionsoff
%  \newpage
%\fi

\clearpage

\section*{References}
%\bibliographystyle{elsarticle-num}
%\bibliography{cviu}

%\vfill


\end{document}